\documentclass{article} 
\usepackage{iclr2023_conference,times}

\usepackage{amsmath,amsfonts,bm}

\def\eqref#1{equation~\ref{#1}}

\def\1{\bm{1}}

\DeclareMathAlphabet{\mathsfit}{\encodingdefault}{\sfdefault}{m}{sl}
\SetMathAlphabet{\mathsfit}{bold}{\encodingdefault}{\sfdefault}{bx}{n}

\DeclareMathOperator*{\argmax}{arg\,max}

\usepackage{hyperref}
\usepackage{url}

\newcommand{\name}{Decomposed Prompting}
\newcommand{\acro}{\textsc{DecomP}\xspace}

\newcommand{\hotpot}{HotpotQA\xspace}

\newcommand{\CoT}{CoT\xspace}
\newcommand{\COT}{CoT\xspace}

\newcommand{\kth}{k^{\mathrm{th}}}

\newcommand{\textbox}[2]{
\begin{center}
\noindent\fbox{%
    \parbox{#1}{%
      {#2}
      }%
    }%
\end{center}
}

\usepackage{booktabs}
\usepackage{graphicx}
\usepackage{wrapfig}
\usepackage{listings}
\usepackage{xcolor}
\usepackage{algpseudocode}
\usepackage{algorithm}
\usepackage{subcaption} 
\usepackage{xspace}

\title{\textsl{\name}: A Modular Approach for Solving Complex Tasks}

\author{%
  Tushar Khot$^{\clubsuit}$, Harsh Trivedi$^{\heartsuit}$, Matthew Finlayson$^{\clubsuit}$, Yao Fu$^{\spadesuit}$\thanks{Work done during internship at Allen Institute for AI}, \\
  \textbf{Kyle Richardson$^{\clubsuit}$, Peter Clark$^{\clubsuit}$, Ashish Sabharwal$^{\clubsuit}$}\\
  $^{\clubsuit}$Allen Institute for AI\quad\quad $^{\heartsuit}$Stony Brook University\quad \quad 
  $^{\spadesuit}$University of Edinburgh\quad \quad 
  \\
  tushark@allenai.org, hjtrivedi@cs.stonybrook.edu, matthewf@allenai.org, yao.fu@ed.ac.uk,\\
  kyler@allenai.org, peterc@allenai.org, ashishs@allenai.org  \\
}

\iclrfinalcopy 
\newcommand{\headertext}{Published as a conference paper at ICLR 2023} 
\begin{document}

\maketitle

\begin{abstract}
Few-shot prompting is a surprisingly powerful way to use Large Language Models (LLMs) to solve various tasks. However, this approach struggles as the task complexity increases or when the individual reasoning steps of the task themselves are hard to learn, especially when embedded in more complex tasks. To address this, we propose \name, a new approach to solve complex tasks by decomposing them (via prompting) into simpler sub-tasks that can be delegated to a shared library of prompting-based LLMs dedicated to these sub-tasks. This modular structure allows each prompt to be optimized for its specific sub-task, further decomposed if necessary, and even easily replaced with more effective prompts, trained models, or symbolic functions if desired.

We show that the flexibility and modularity of \name\ allows it to outperform prior work on few-shot prompting using GPT-3. On symbolic reasoning tasks, we can further decompose sub-tasks that are hard for LLMs into even simpler solvable sub-tasks. When the complexity comes from the input length, we can  recursively decompose the task into the same task but with smaller inputs. We also evaluate our approach on textual multi-step reasoning tasks: on long-context multi-hop QA, we can more effectively teach the sub-tasks via our separate sub-tasks prompts; and on open-domain multi-hop QA, we can easily incorporate a symbolic information retrieval module within our decomposition framework, leading to improved performance on both tasks.\footnote{Datasets, Code and Prompts available at \url{https://github.com/allenai/DecomP}.}
\end{abstract}

\section{Introduction}
Large Language Models (LLMs) such as GPT-3 \citep{gpt3} have been shown to solve various tasks given only a few examples as prompts, also referred to as in-context learning. These models can even perform more complex reasoning tasks when shown the sequence of simple reasoning steps needed to perform the complex task as a prompt~\citep{Wei2022ChainOT,Nye2021ShowYW}. In essence, the sequence of reasoning steps, such as in Chains-of-Thought (\COT) prompting \citep{Wei2022ChainOT}, demonstrates how to decompose the complex task as well as how each reasoning step should be performed. However, as tasks become more complex, few demonstrations of the complex task aren't sufficient for current models to learn to perform all necessary reasoning steps. E.g., few-shot demonstrations of concatenating the $\kth$ letter of words in a string is insufficient for GPT-3 to learn to extract the $\kth$ letter, or learn to answer hard single-hop questions when only provided a few demonstrations of multi-hop questions. Additionally, it is unclear whether tasks such as document retrieval and integration, for knowledge-intensive tasks, can even be done by few-shot prompts. 

\begin{figure}[htbp]
    \centering
    \includegraphics[width=\linewidth]{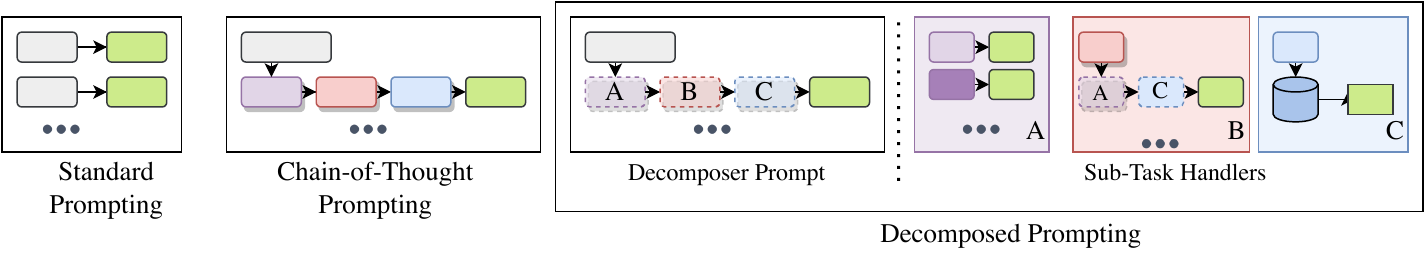}
    \caption{While standard approaches only provide labeled examples (shown as a grey input box with green label box), Chain-of-Thought prompting also describes the reasoning steps to arrive at the answer for every example in the prompt. \name, on the other hand, uses the decomposer prompt to only describe the procedure to solve the complex tasks using certain sub-tasks. Each sub-task, indicated here with A, B and C is handled by sub-task specific handlers which can vary from a standard prompt (sub-task A), a further decomposed prompt (sub-task B) or a symbolic function such as retrieval (sub-task C)
    }
    \label{fig:intro}
\end{figure}
   
To address these limitations, we propose \textbf{\name} (\acro), a new approach to solve complex tasks by instead decomposing them into simpler sub-tasks and delegating these to sub-task specific LLMs, with both the decomposer and the sub-task LLMs (henceforth, \emph{sub-task handlers}) having their own few-shot prompts. Fig~\ref{fig:intro} illustrates our approach. The decomposer prompt only describes a sequence of sub-tasks (A, B, and C) needed to solve the complex tasks, indicated with the dashed lines. Each sub-task is then delegated to the corresponding sub-task handler shown on the right.

Using a software engineering analogy, the decomposer defines the top-level \emph{program} for the complex task using interfaces to simpler, sub-task functions. The sub-task handlers serve as modular, debuggable, and upgradable \emph{implementations} of these simpler functions, akin to a software library. If a particular sub-task handler, say the one for identifying the $\kth$ letter or retrieving a document, is not performing well enough, we can debug this handler in isolation, explore alternative prompts or implementations, and seamlessly plug the improved module back into the overall system, as a systematic way to try to improve performance on the complex end-task. 

This approach has several advantages over prior work (as also shown in the figure). The sub-task handlers can be shown a broader and richer set of examples (of the simpler task) than the specific ones needed for the complex task prompt (task A). If a sub-task is too complex, it can be further decomposed into simpler sub-tasks (task B). Similar to software libraries, these sub-task handlers can be shared across multiple tasks; e.g., here tasks A and C are reused in the model for task B. As noted above, a sub-task handler can be easily swapped with an improved implementation without any change to the rest of the system. Few-shot prompt based LLMs can be even replaced with a symbolic system for tasks more suited for non-neural methods; e.g., task C uses a symbolic retrieval system such as Elasticsearch that can handle very large-scale corpora. Lastly, we can even improve upon prior work by simply adding an \emph{error-correcting} sub-task handler as a post-processing step.

To illustrate these advantages of \acro, we empirically evaluate it against prior work on eight challenging datasets using GPT3 models: (1) On a task of concatenating the $\kth$ letter, we show that our approach of factoring out each sub-task allows us to more effectively teach the sub-problem of extracting the $\kth$ letter(specifically, by decomposing it into even easier sub-tasks). (2) On a task of reversing a list, we show that \acro\ allows us to extend the capabilities of a weaker model and build a scale-invariant system by recursively decomposing the task into reversal of smaller and smaller lists. (3) On a task of long-context QA~\citep{Khot2022HeyAC}, our approach allows each sub-task handler to accommodate more examples than feasible with \COT prompting leading to better QA performance. (4) On three multi-hop open-domain QA datasets~\citep{hotpotqa,xanh2020_2wikimultihop,musique}, we can incorporate a symbolic retrieval (ElasticSearch) API as the handler for the retrieval sub-task leading to better results than \COT. (5) On two Math QA datasets~\citep{cobbe2021gsm8k,roy-roth-2015-solving}, we can post-process CoT to easily fix frequent formatting errors, resulting in a surprisingly high improvement of 14-17 pts.

\section{Related Work}
\paragraph{Few-shot Prompts for Multi-Step Reasoning}

Large-scale Language models (LLMs) have been shown to learn various NLP tasks given just few examples as prompts~\citep{gpt3}. Recently, they have also been successfully applied to various multi-step reasoning tasks by providing the intermediate reasoning steps, i.e. Chain-of-Thought~\citep{Wei2022ChainOT,chowdhery2022palm}, needed to arrive at the answer. An alternate approach has been to compose multiple LLMs or LLMs with symbolic functions to perform multi-step reasoning~\cite[inter alia]{jung2022maieutic,creswell2022selection,selfask,talm,pal,Schick2023ToolformerLM}. We view these prior works as specialized systems with a pre-defined decomposition structure.  

The closest works to our approach are the ideas of least-to-most prompting~\citep{Zhou2022LeasttoMostPE} and successive prompting~\citep{succprompting} where one prompt/model is used to generate the sub-questions needed to answer a complex question and a second prompt/model sequentially answers these sub-questions. In contrast, our approach allows for diverse decomposition structures including recursion and other non-linear decomposition structures. E.g., by definition, least-to-most asks questions from easiest to the hardest and requires an LLM to eventually answer the complete question (“most” in least-to-most) whereas we have no such restriction. Additionally, we iteratively generate new questions based on previous answers (similar to successive prompting) and can explicitly assign different prompts or symbolic systems to answer each sub-question.

\paragraph{Modular Approaches for Multi-Step Reasoning}

Our work follows a long literature in NLP on neural modular modeling architectures \citep{andreas2016neural,talmor2018web,min-etal-2019-multi,jiang2019self,gupta2020neural,perez2020unsupervised,khot-etal-2021-text,levine2022standing} for question-answering and other tasks. We take particular inspiration from the \emph{Text Modular Networks} approach of \citet{khot-etal-2021-text}, whereby problem decomposition consists of a learned \emph{next question} generator trained to generate questions in the language of a collection of textual and symbolic agents. Best-first search strategy was used to explore the space of possible decompositions during inference. In contrast to this work, which largely centered around supervised training of the next-question generator \emph{given existing agents}, we leverage the power and recent successes of few-shot LLMs to build both the decomposer and the sub-task agents that best fit the ideal decomposition. This has the advantage of obviating the need for specialized supervised training data that may not always be available for all sub-tasks -- a key bottleneck of this prior work.

\section{\name}

As with conventional \emph{few-shot} prompting, the goal is to teach an LLM to find an answer $A$ to a query $Q$ using a small set of \emph{in-context} examples $D = \{E_{1},...,E_{|D|}\}$. The answer $A$ is obtained from the underlying distribution $p(A \mid Q,D,\theta)$ \citep{dohan2022language}. In the most basic few-shot setup, examples take the form $E_j = (Q_j, A_j)$. In the case of \COT-style prompting, the goal is to obtain answers by first generating a sequence or chain of intermediate reasoning steps or ``thoughts'' $T$, and then deriving the final answer based on $T$. To teach this ability, one uses more sophisticated in-context examples that take the form $E_{j} = (Q_{j},(T_{j,1},\ldots,T_{j,k}),A_{j})$.

In \acro, the core is a \emph{decomposer} LLM that tries to solve a complex task by generating a \textbf{prompting program} $P$ for it. Each step of $P$ directs a simpler sub-query to a function in an auxiliary set of \textbf{sub-task functions} $\mathcal{F}$ available to the system. Given a query $Q$ whose answer is $A$, the program $P$ is a sequence of the form $\big((f_1,Q_1,A_1),...,(f_k,Q_k,A_k)\big)$ where $A_k$ is the final answer predicted by $P$ and $Q_i$ is a sub-query directed to the sub-task function $f_i \in \mathcal{F}$. $P$ is executed by a high-level imperative \textbf{controller}, which passes the inputs and outputs between the decomposer and sub-task handler until a stopping condition in $P$ is met and the final output obtained.

To teach the decomposer LLM in a few-shot prompting manner, we use in-context examples that take the form $E_j = \big( (Q_j, \big(f_{j,1},Q_{j,1},A_{j,1}),...,(f_{j,k_j},Q_{j,k_j},A_{j,k_j})\big) \big)$ where $A_{j,k_j} = A_j$ is the final answer for $Q_j$ and $(Q_{j,1}, \ldots, Q_{j,k_j})$ is a decomposition of $Q_j$. Each sub-task function $f$, in turn, is operationalized via a sub-task handler as an in-context prompting LLM (e.g., a separate \COT-style prompt or a additional prompting program dedicated to that sub-task), or any other symbolic or learned function (e.g., a calculator or specialized supervised trained model).

\subsection{Decomposed Prompts}

\begin{figure}[t] 
    \centering
    \includegraphics[width=0.8\linewidth]{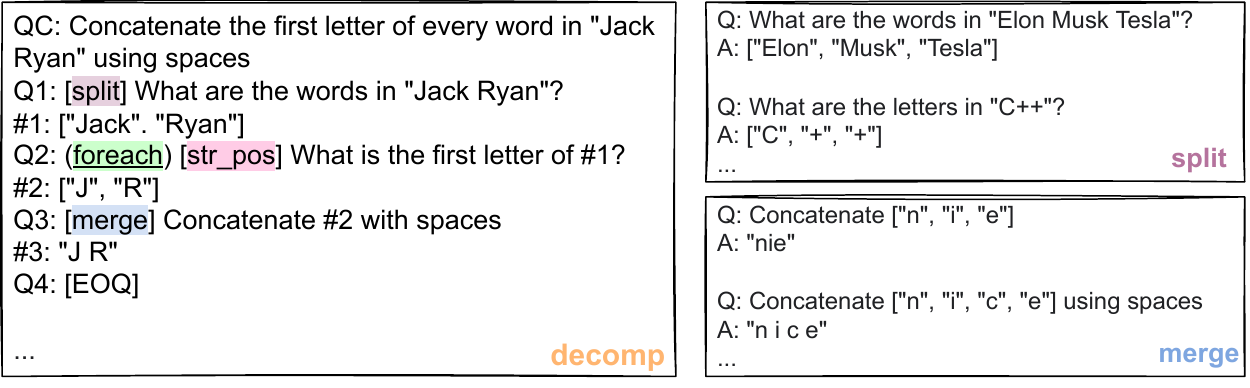}
    \caption{Prompts for the decomposer and the \texttt{split} and \texttt{merge} sub-tasks used by the decomposer. The decomposer specifies the sequence of questions and corresponding sub-tasks (within square braces). The sub-task prompts can be written independent of the complex task examples and can even capture generalizations, e.g., letters in word (\texttt{split}) and no delimiter (\texttt{merge}).}
    \label{fig:letter_cat_prompts}
\end{figure}

To illustrate this with an example, consider a multi-step task such as ``Concatenate the first letter of every word in $str$ using a space''. We can solve this task by decomposing it into a sequence of three simple sub-tasks: 1) Collect the list of words in the $str$; 2) For each word, extract the third letter; 3) Concatenate the extracted letters using space as the separator. Fig.~\ref{fig:letter_cat_prompts} shows an example decomposition prompt for this task. Much like a conventional structured program, the top-level \texttt{decomp} prompt provides an example program $E_j$ using three sub-task functions: $f_1:$\texttt{split} that \emph{splits words in an input string}, $f_2:$\texttt{str\_pos} that \emph{finds character positions in strings} and $f_3:$\texttt{merge} that \emph{concatenates characters}. In this case, we operationalize each sub-task function as a separate in-context prompt (e.g., using a standard prompting approach for \texttt{split} and \texttt{merge} on the right side), each containing a set of in-context examples that are independent of the original complex task. 

In addition to the three functions described above, additional control structure is included, such as the symbolic function \texttt{foreach}, which iterates over arrays and references to previous answers such as \#1. We note that such a helper function is not strictly necessary (e.g., we could directly generate ``Q2': What is the first letter of Jack?'' and ``Q3': What is the first letter of Ryan?'' instead of Q2 in the figure) and is added to reduce the manual effort needed to specify the decomposition and also reduce potential errors during decomposition. In our experiments we use two of the compositional operators defined by \citet{Khot2022HeyAC} (see appendix for details), although it is capable of using all their operators (which also capture the QDMR operators from \citet{Wolfson2020Break}).

\begin{figure}[htbp]
    \centering
    \includegraphics[width=0.9\linewidth]{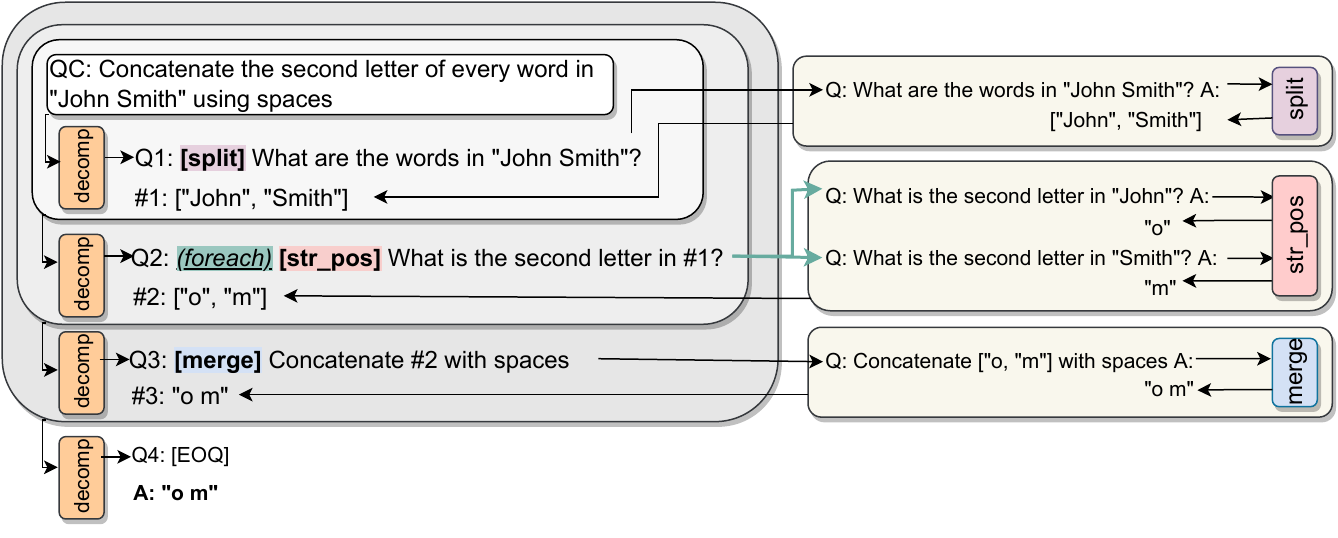}
    \caption{The inference procedure in \acro\ iteratively calls the decomposer prompt to generate the next question and sub-task at each step, given the current history of question and answers. The generated question is then routed to the assigned sub-task handler (with some handling of special operators, when needed). When the special end-of-questions \texttt{[EOQ]} marker is generated, the previous answer is returned as the final prediction.}
    \label{fig:letter_exec}
\end{figure}
\subsection{Prompt Execution and Inference}

Given a new question and a set of background in-context examples $D$, the inference (i.e., the program construction and execution) process is illustrated in Fig.~\ref{fig:letter_exec}. The new complex question is fed to the decomposer prompt to get the first sub-question to be asked to the \texttt{split} prompt. With the help of our symbolic controller, the answer generated from this prompt is then appended to the decomposer prompt to get the second sub-question, $Q2$. Due to the \texttt{foreach} operator in the generated question, $Q2$ results in two questions (one for each word in $\#1$) to be fed to the \texttt{str\_pos} prompt. The answers are combined into an array to get the answer $\#2$. The entire decomposition history is used to generate $Q3$  and passed to the \texttt{merge} prompt to get the final answer. Since the task has been solved, the decomposition prompt produces the special end-of-sequence marker([EOQ]) and the last answer is returned as the final answer. Formally, performing inference involves finding the best answer $A$ to a new query $Q$, which in the simplest form involves computing the MAP answer using the LLMs predictive distribution for $A$, i.e., $\hat{A} = \argmax_{A} p(A \mid D,Q,\theta)$ \citep{dohan2022language}. For practicality, such computations are approximated using greedy search in our experiments.

\subsection{\acro\ Capabilities}

\begin{figure}[htbp]
    \centering
    \includegraphics[width=\linewidth]{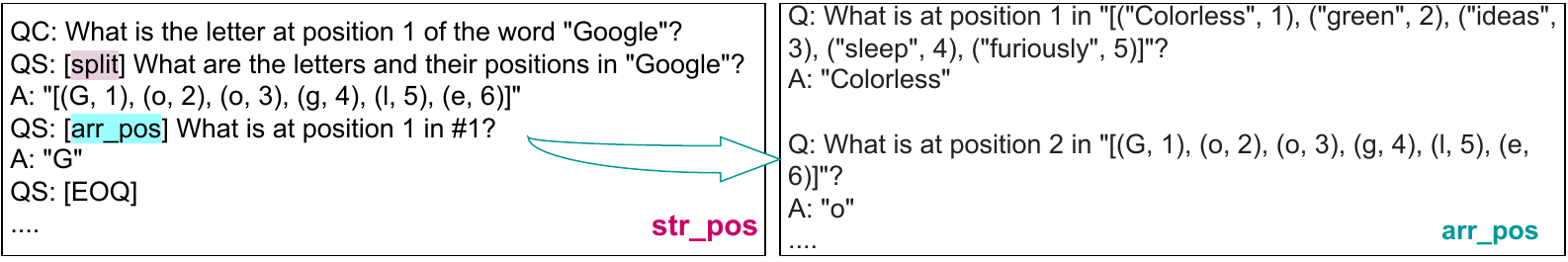}
    \caption{Since identifying the $k^{th}$ character is challenging for GPT3 \texttt{davinci-002} model, we further decompose it into two simpler sub-tasks: split the word into its letters (using the shared sub-task \texttt{split}) and then return the $k^{th}$ item of this list using the \texttt{arr\_pos} prompt.}
    \label{fig:str_pos_prompts}
\end{figure}

\textbf{Hierarchical Decomposition}
Certain sub-tasks, even when given many examples, are not solvable with few-shot prompting. E.g., we found identifying the $\kth$ letter of a string to be challenging for the GPT3 \texttt{text-davinci-002} model. In such a scenario, we can decompose the sub-task prompt further, to first identify the letters and their position and then select the $\kth$ element of this array (see Fig.~\ref{fig:str_pos_prompts}). We can also re-use existing sub-task prompts in our framework. E.g., the \texttt{split} prompt can be reused since it was developed for the general task of splitting strings.\footnote{Appendix~\ref{app:prompts} contains the complete \texttt{split} prompt which has examples for questions such as ``Q: What are the letters and their positions in ``Mathison"?''.}

\begin{figure}[htbp]
    \centering
    \includegraphics[width=\linewidth]{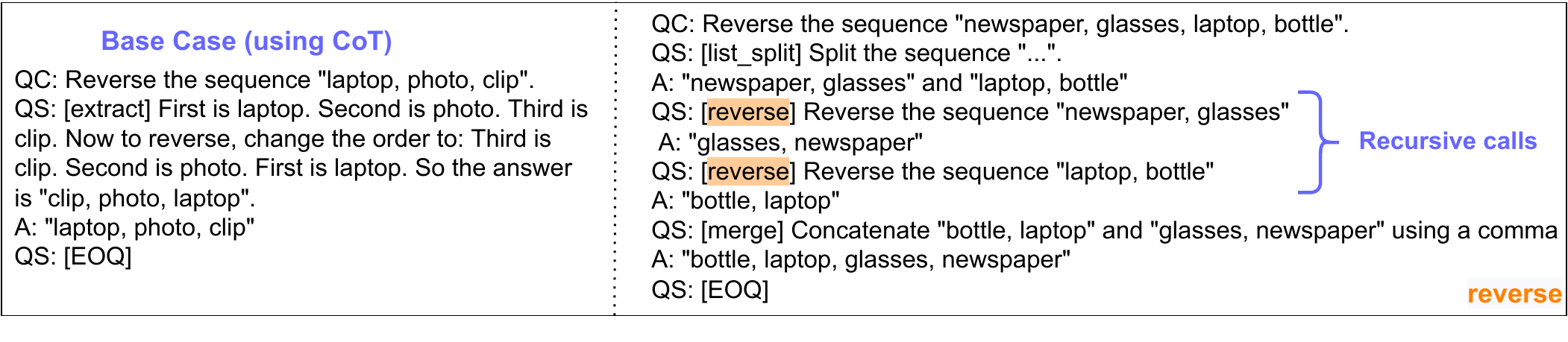}
    \caption{Sample prompt for recursive decomposition for reversing lists. Each list is split into two halves and each half is reversed and concatenated in the reverse order. We can recursively split a list till we hit the base case (lists of length 3 here) where existing approaches such as CoT are accurate.}
    \label{fig:reverse_prompts}
\end{figure}

\textbf{Recursive Decomposition}
Some problems can be naturally broken down into one or more smaller problems of the same form.
Recursive algorithms such as merge sort use this idea to solve large problems efficiently, using a succinctly described method.
We apply this same principle in \acro by allowing the decomposer prompt to recursively call itself, as shown in Fig.~\ref{fig:reverse_prompts} for the task of list reversal. By using recursion, we are able to generalize any base prompting approach (CoT in this figure) to much longer lists by breaking the input into smaller and smaller lists till we reach a list length where the model is highly accurate. Such recursive approaches can not be described by current methods such as \COT and standard prompting. Least-to-most prompting~\citep{Zhou2022LeasttoMostPE} also proposes a similar solution but differs in two key aspects (a) it has to identify all the sub-problems in one-shot instead of our iterative top-down decomposition (b) it has to learn to identify the relevant answers from the previous solutions which we get for free from our decomposition.

\paragraph{External API Calls}

\begin{figure}[htbp]
    \centering
    \includegraphics[width=\linewidth]{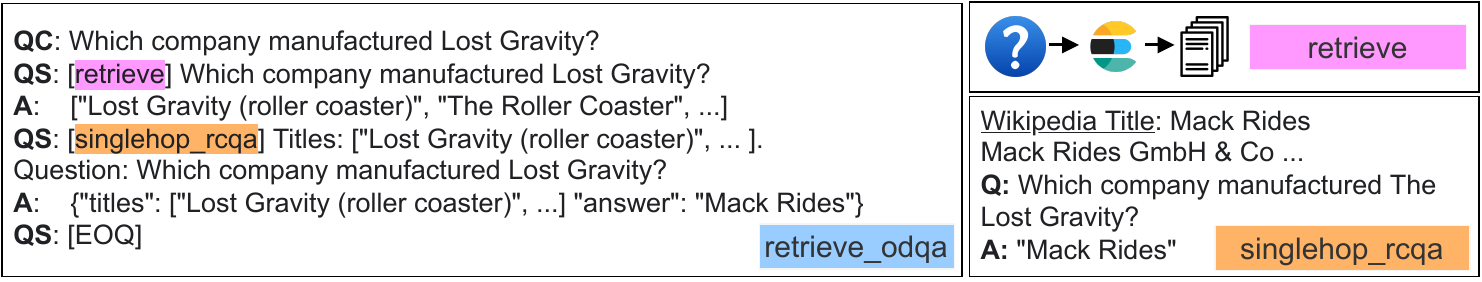}
    \caption{A Decomposed Prompt to answer open-domain questions using Elasticsearch-based retrieval. Full usage of this prompt for open-domain multihop questions is given in Fig. ~\ref{fig:hotpotqa_prompt_detailed}.}
    \label{fig:hotpotqa_prompt_intro}
\end{figure}

In certain cases, the sub-tasks may not be feasible to solve using only a LLM. E.g., retrieving knowledge from a KB or large corpus. Such sub-tasks, however, can be easily solved using existing systems such as retrieving documents using an Elasticsearch index or webpages using Google search~\citep{Lazaridou2022InternetaugmentedLM}.  Fig.~\ref{fig:hotpotqa_prompt_intro}  shows how \acro\ can easily use such a system to retrieve the relevant documents and answer a single-hop open-domain question.

\section{Case Studies}
We showcase \acro's strengths through four tasks; two symbolic manipulation tasks similar to those investigated by \citet{Wei2022ChainOT} and two existing textual multi-hop reasoning tasks. Unless specified, we use \texttt{text-davinci-002} InstructGPT3 model~\citep{Ouyang2022TrainingLM} as the LLM and report the Exact Match (EM) numbers, following prior work. For order-independent list answers, we evaluate set equality as EM. We compare our approach to \COT rather than each specific decomposition structure used in prior work. See App.~\ref{app:prompts} for the complete prompts for all our tasks.

\subsection[k-th letter concatenation (Hierarchical Decomposition)]{$\kth$ letter concatenation (Hierarchical Decomposition)}
We compare \acro to \CoT prompting for concatenating letters at the $\kth$ position. All prompts contain examples of concatenating letters in position 1, 4, and last position of strings with 3 words. We create three different prompts for all our baselines and present the average to account for variance due to the choice of examples following \citet{Perez2021TrueFL}. We use the \texttt{decomp}, \texttt{split}, \texttt{str\_pos} (further decomposed as shown in Fig.~\ref{fig:str_pos_prompts}), and \texttt{merge} prompts for decomposition prompting. We adapt the \COT for last letter concatenation from prior work~\citep{Wei2022ChainOT} for this task as shown below. In addition, we consider a \emph{rolled out} version of our decomposition prompts in terms of a \COT, i.e., we describe the entire decomposition process (identify words, split each word into letters, take $\kth$ letter and concatenate) as a single \COT. e.g, for the question ``Take the letters at position 4 of the words in "Herbert Alexander Simon" and concatenate them using a space.'', we use the \COT:

\noindent \begin{minipage}{0.5\textwidth}
\textbox{19em}{\textbf{Chain-Of-Thought}\\
\small
The letter at position 4 of "Herbert" is "b". The letter at position 4 of "Alexander" is "x". The letter at position 4 of "Simon" is "o". Concatenating "b", "x", "o" using a space leads to "b x o". So, "Herbert Alexander Simon" outputs "b x o".
...}
\end{minipage}
\begin{minipage}{0.5\textwidth}
\textbox{19em}{\textbf{Chain-Of-Thought (rolled out)}\\
\small
The words in "Herbert Alexander Simon" are "Herbert", "Alexander", and "Simon". The letters and their positions in "Herbert" are "[(H, 1), (e, 2), (r, 3), (b, 4), (e, 5), (r, 6), (t, 7)]". The letter at position 4 in this sequence is "b". $\cdots$ outputs "b x o".
...}
\end{minipage}

\begin{figure}
    \begin{minipage}{.475\textwidth}
    \includegraphics[width=\linewidth]{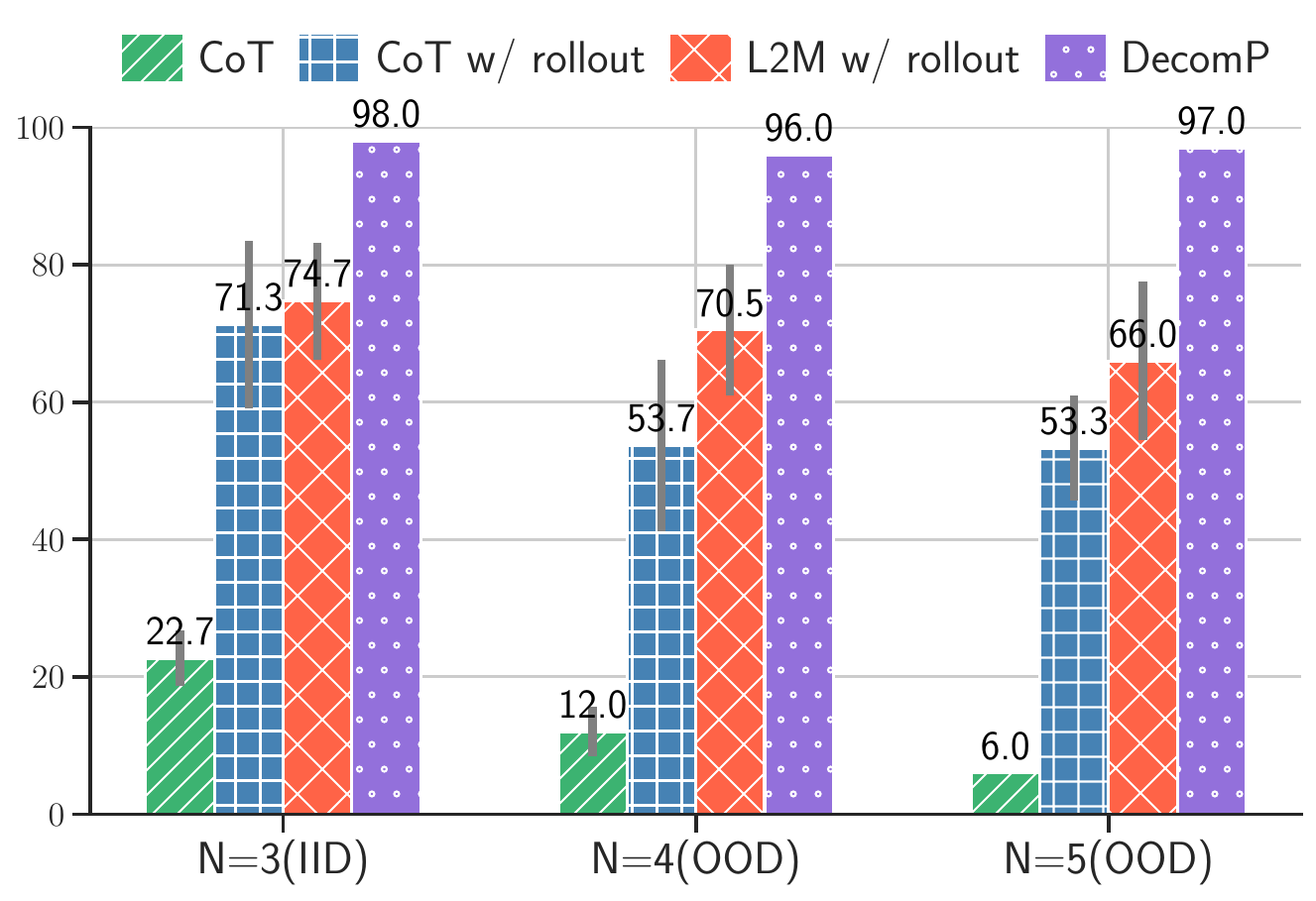}
        \caption{EM Results on the $k^{th}$ letter concatenation task (k=3) using space as delimiter with different number of words in the input. \acro\ outperforms and generalizes better than \COT as well as Least-to-most prompting.
    }
    \label{fig:letter_cat_space_results}
    \end{minipage}
    \hfill
    \begin{minipage}{.475\textwidth}
    \includegraphics[width=\linewidth]{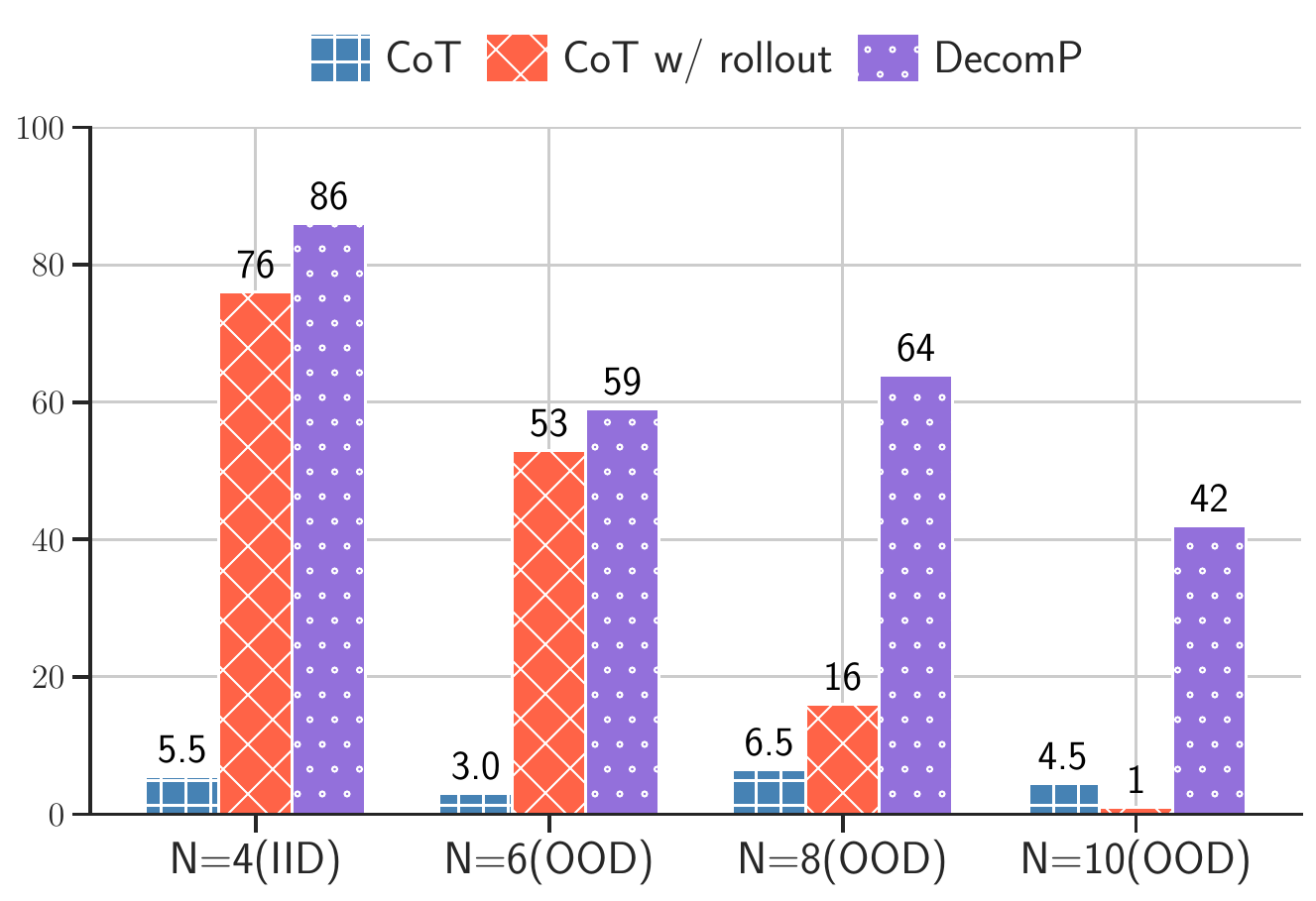}
\caption{EM results on reversing sequences. Incorporating \CoT in \acro greatly increases the ability of the model to generalize to new sequence lengths.}\label{fig:reverse}   \end{minipage}
\end{figure}

We similarly adapt the least-to-most prompt~\citep{Zhou2022LeasttoMostPE} to include rollout. (see App.~\ref{app:prompts}). We compare these four prompting techniques on 4 datasets to evaluate generalization along 3 axes: (1) new letter position $k=3$;\footnote{Note that none of the sub-task prompts contain examples for this position.} (2) longer inputs, \#words=4 and 5; (3) new delimiter ";". The words in the test examples come from a list of most popular first and last names.\footnote{\url{forebears.io/earth/forenames} and \url{forebears.io/earth/surnames}} All evaluation datasets have 100 examples. We present results on space as a delimiter averaged across three prompts in Fig.~\ref{fig:letter_cat_space_results}.\footnote{We obtain similar results with semi-colon shown in Fig.~\ref{fig:letter_cat_semic_results} in the appendix.}

\textbf{\acro outperforms chain-of-thought and least-to-most prompting}, even when the prompt uses the same reasoning procedure as the rolled out decomposition. This shows that the separate prompts are more effective at teaching hard sub-tasks than a single \COT prompt. 

\textbf{\acro generalizes perfectly to longer sequences.} As the length of the input sequence increases, our approach continues to achieve close to 100\% accuracy on this task.\footnote{Note that we report aggregate metrics for \acro\ too but the std. dev is zero here}  
The \COT-based approaches drop noticeably in their scores with longer input lengths, widening the performance gap.

\subsection{List Reversal (Recursive Decomposition)}

We use the task of reversing lists of words\footnotemark 
to show how recursive \acro
enables length generalization.
We adapt the relevant \COT prompt from \citet{Wei2022ChainOT},
and integrate it in a decomposed prompt.
As a control, we also compare to a \COT version w/ rollout of our decomposed prompt. 
All prompts contain the same 3 examples of reversing word sequences 
with 3-5 items. 
We evaluate all prompts for generalization 
to 4, 6, 8, and 10-item sequences.
Here we use \texttt{davinci-001} to show that \acro enables a weaker model approach \texttt{davinci-002}'s performance (which does solve this task). \footnotetext{We use the vocabulary from \citet{Wei2022ChainOT}: \url{https://www.vocabulary.com/lists/189583}} We use the strategy from Fig.~\ref{fig:reverse_prompts} and provide our prompts in App.~\ref{app:prompts}. Fig.~\ref{fig:reverse} shows the results of the prompting strategies on different input lengths.

\textbf{\acro improves the length generalization of few-shot prompting.} While our base CoT prompt does not generalize at all to longer sequences, our approach can recursively decompose the problem and achieve better length generalization. 
Moreover, the \COT version of our decomposition strategy fails 
because the unrolled prompt becomes too long and convoluted 
without the ability to abstract away sub-modules. 

\subsection{Long-Context Question Answering}

We next evaluate on the CommaQA-E dataset \citep{Khot2022HeyAC} under the reading comprehension setting. The dataset consists of synthetically generated entities (e.g. Erowid award), facts (``Wetherality was an actor in the movie Dewbar.'') and multi-hop questions (e.g., ``What awards have the actors of the Erowid winning movies received?''). Due to the presence of many distractors and, as a result, longer context, this dataset has been shown to be hard for standard LMs even when fine-tuned. 

\begin{figure}[bhtp]
    \centering
    \includegraphics[width=\linewidth]{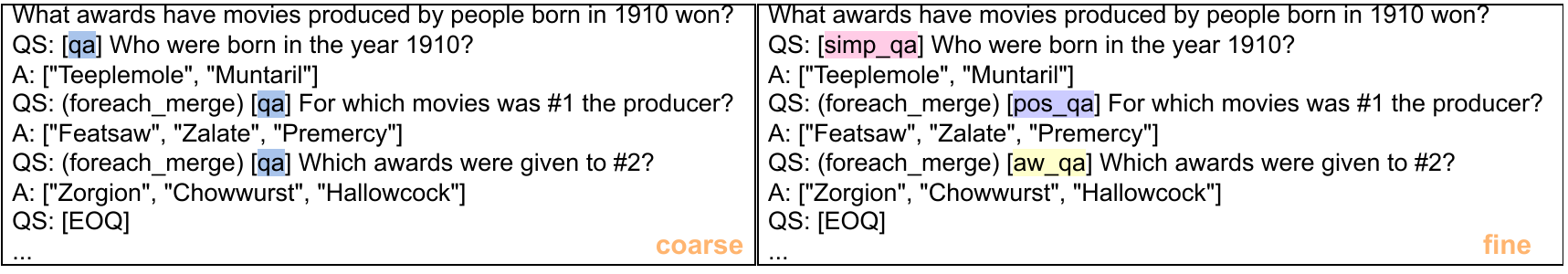}
    \caption{Sample prompts used for the CommaQA dataset. On the left, the coarse-grained decomposition defines a single QA sub-task with all single-hop questions being delegated to a single sub-task handler. On the right, the fine-grained decomposition assigns questions to three different sub-tasks (see App.~\ref{app:prompts} for their prompts) depending on the question type. This allows us to provide more examples for each question type allowing the model to learn the sub-task more effectively.}
    \label{fig:commaqa_prompts}
\end{figure}

To fit these questions within GPT3's context limit (2049 tokens), we generate a smaller version of the CommaQA-E dataset and of the compositional generalization split such that we can fit at least four examples in the context for \COT prompts. The \COT prompts describe the sequence of facts needed to arrive at the answer (see App.~\ref{app:prompts} for all the prompts).

\begin{wrapfigure}{r}{0.45\textwidth}
\begin{center}
     \includegraphics[width=\linewidth,trim=0cm 0cm 0cm 0.5cm, clip]{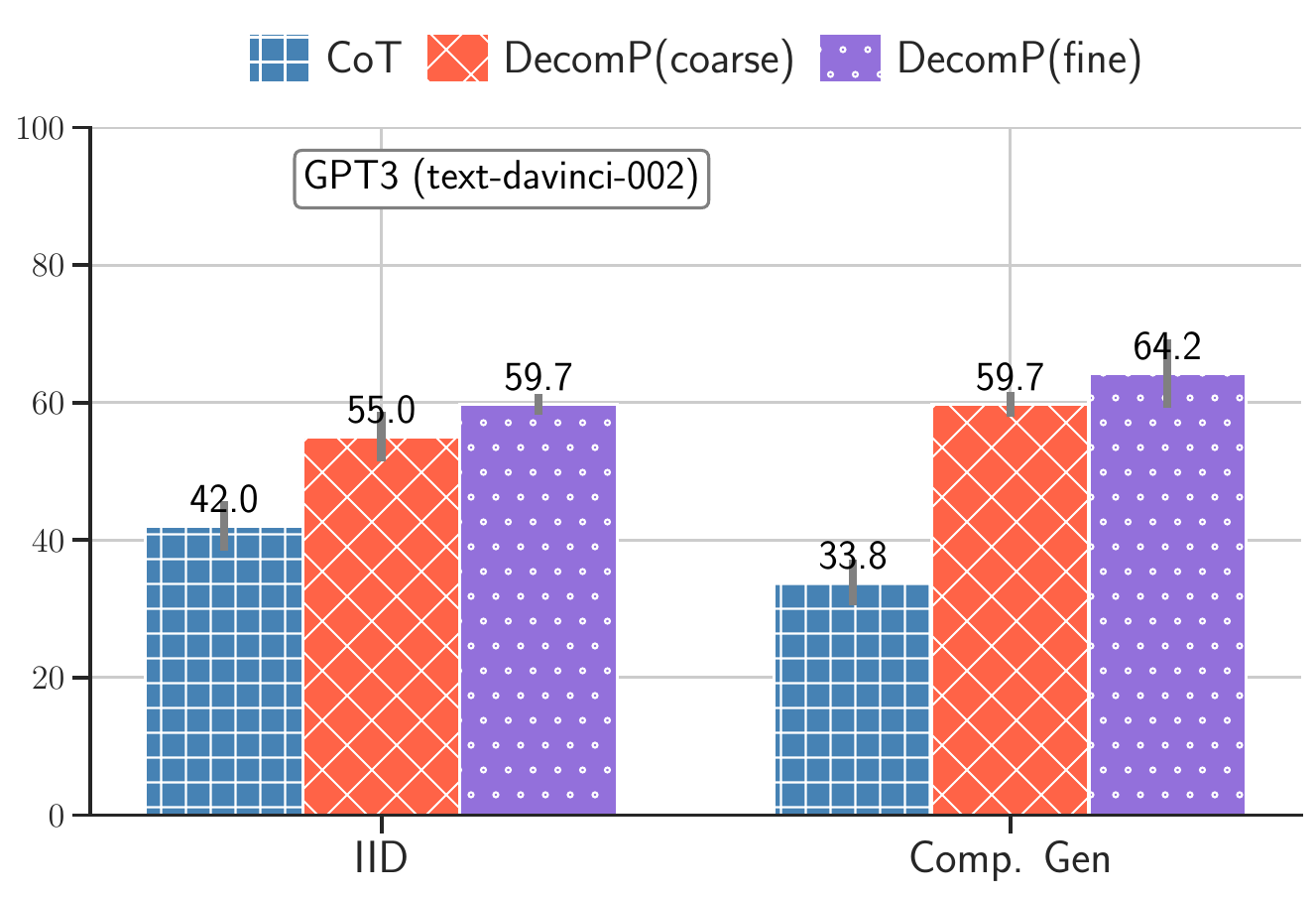}
  \end{center}
\caption{EM results on the CommaQA-E datasets. \acro\ always outperforms \COT, with fine-grained marginally out-performing coarse-grained decomposition.}\label{fig:commaqa_results}\end{wrapfigure}

For \acro, we can separate the task of decomposition (independent of the context) from the sub-tasks of single-hop question answering. As shown in Fig.~\ref{fig:commaqa_prompts}, we provide examples of the context-independent decomposition in the decomposer prompt and use the separate sub-task prompts to teach the QA skill over the given context. Additionally, we can choose the granularity of decomposition to trade off human effort for increased accuracy. For example, we could have single QA prompt to handle all the questions or create QA prompts for different classes of questions. In our experiments, each sub-task prompt contains 8 QA examples (2 questions/para). We evaluate three different prompts and report the average results in Fig.~\ref{fig:commaqa_results}.

We make three observations on CommaQA. \textbf{\acro is more accurate than \COT} irrespective of the granularity of decomposition or the evaluation split.  
\textbf{Finer grained decomposition can help improve task performance} by providing more examples for each class of questions, which in turn increases single-hop QA accuracy.
 \textbf{\acro generalizes to new compositions} such as the compositional generalization split of CommaQA, which tests models on unseen compositions of relations observed in the training set. While CoT has a drop in score, both decomposition-based approaches actually get a small bump (the subset of relations used in this split are easier for our QA models).

\subsection{Open-Domain Question Answering}
\label{subsec:hotpot}

Next, we demonstrate the ability of our approach to integrate external API calls on the task of open-domain multihop question answering. We evaluate our approach on three datasets: (1) 2WikiMultihopQA~\citep{xanh2020_2wikimultihop} (2) MuSiQue~\citep{musique} (3) \hotpot~\citep{hotpotqa}. We describe the open-domain versions of these datasets in more detail in App.~\ref{app:odqa} We use the Codex (code-davinci-002) model here since it can fit the much longer contexts needed. We also evaluate the impact of model scale on \acro\ by using models from the Flan-T5 family: Flan-T5-Large (0.7B), Flan-T5-XL (3B), and Flan-T5-XXL (11B).\footnote{We still use GPT3-sized models for decomposition since only these models are reliably able to produce the required structured outputs.}
\begin{figure}[htbp]
    \centering
    \includegraphics[width=\linewidth]{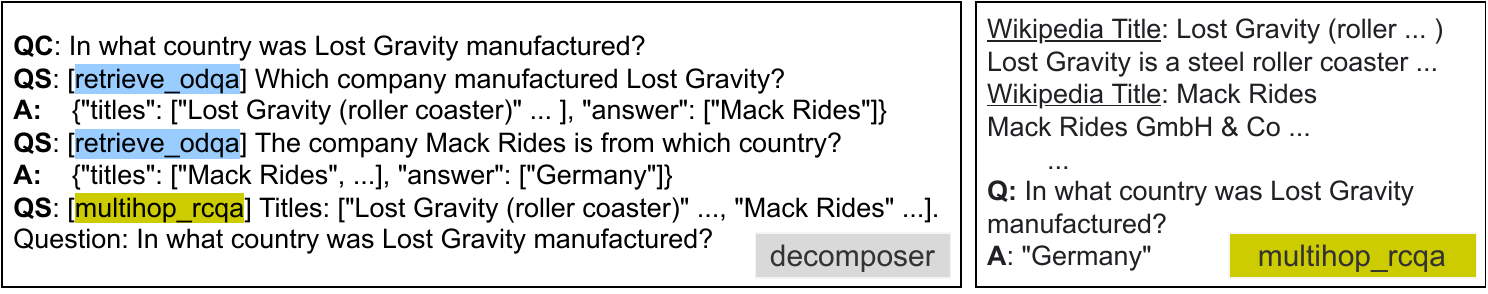}
    \caption{The prompt used to answer open-domain multihop questions using Elasticsearch-based retrieval. The \texttt{retrieve\_odqa} prompt is given in Fig.~\ref{fig:hotpotqa_prompt_intro}.}
    \label{fig:hotpotqa_prompt_detailed}
\end{figure}

Fig.~\ref{fig:hotpotqa_prompt_detailed} shows the decomposition prompt we use. The decomposer generates (singlehop) sub-questions and delegates them to \texttt{retrieve\_odqa} (described in Fig.~\ref{fig:hotpotqa_prompt_intro}). As we showed earlier, this module retrieves relevant documents then uses an RC model to answer. \texttt{retrieve\_odqa} returns both the answer and the documents, allowing subsequent sub-questions to use the answers (e.g. ``Mack Rides'') and the \texttt{multihop\_rcqa} model to use the documents. The final \texttt{multihop\_rcqa} model is prompted to produce the answer directly or using CoT given K paragraphs.

We compare our approach against two baselines: \textbf{A. No Context (No-Ctxt),} A closed-book setting baseline where the model must rely only on its parametric knowledge. \textbf{B. NoDecomp Context (\textbf{NoDecomp-Ctxt}),} A simple retrieval baseline where we retrieve K paragraphs using the multi-hop question as the input and use that as context. For both NoDecomp-Ctxt and Decomp-Ctxt, K is selected by hyperparameter tuning (App.~\ref{app:odqa}). We manually annotate CoTs and decompositions for 20 training set questions, and sample 3 prompts of 15 questions each for all approaches. The detailed prompts are given in the Appendix \ref{app:prompts}. We evaluate on 300 held-out dev questions in each dataset.

\begin{figure}[htbp]
    \centering
    \begin{subfigure}{0.48\textwidth}
    \includegraphics[width=\linewidth]{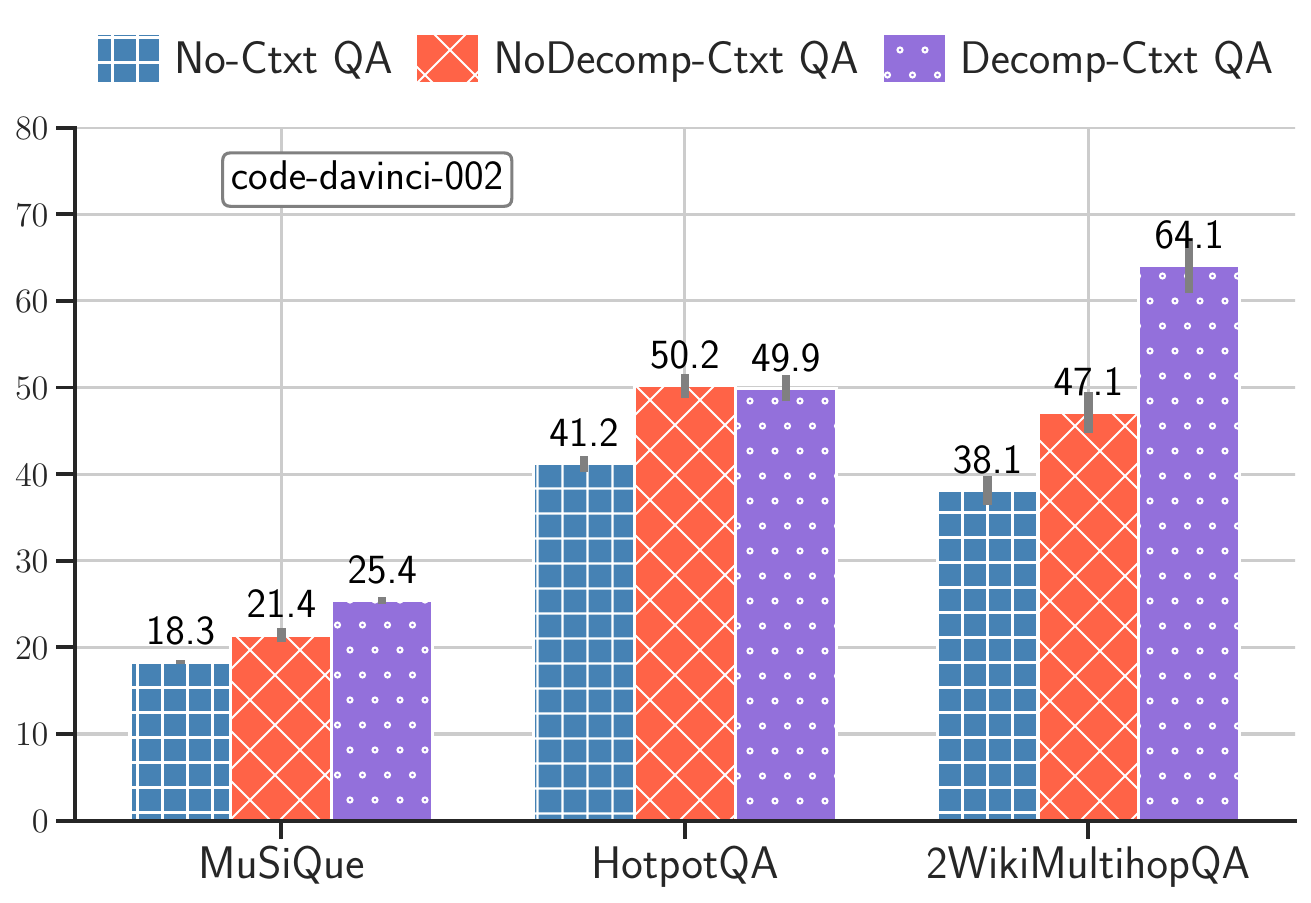}
    \end{subfigure}
    \hfill
    \begin{subfigure}{0.48\textwidth}
    \includegraphics[width=\linewidth]{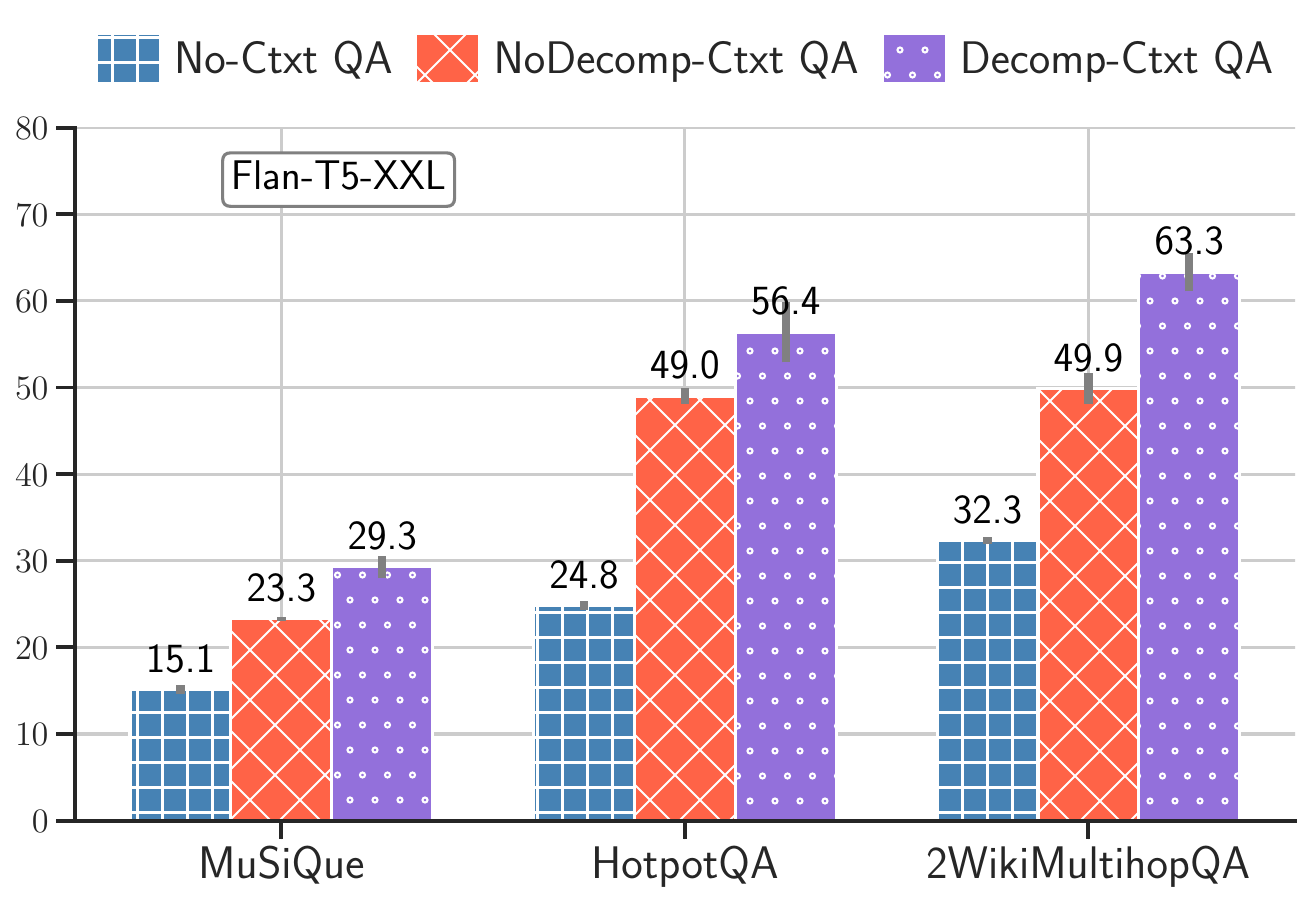}
    \end{subfigure}
    \hfill
    \caption{Answer F1\protect\footnotemark on three open-domain QA datasets using two base LMs: Codex (left) and Flan-T5-XXL (right) with direct prompting. Decomp-Ctxt models (ours) significantly outperforms the No-Ctxt models (no retrieval) in all settings and also outperforms our strong retrieval baseline (NoDecomp-Ctxt QA), with the exception of Codex on HotpotQA where it is comparable. See App. \ref{app:odqa_results} for results on smaller Flan-T5 models and CoT prompting.}
    \label{fig:odqa-results}
\end{figure}
\footnotetext{Answer F1 is computed by treating prediction and ground truth answer as bags of tokens and computing their precision and recall~\citep{rajpurkar-etal-2016-squad}. See HotpotQA~\citep{hotpotqa} for details.}

We present results on all three datasets with direct QA prompts in Fig.~\ref{fig:odqa-results} with other results in App.~\ref{app:odqa}. The Decomp-Ctxt models performs significantly better than No-Ctxt models in all the settings showing that external knowledge can be leveraged to improve few-shot models on open-domain mulithop QA. Furthermore, we show that our Decomp-Ctxt models outperform the strong retrieval baseline (NoDecomp-Ctxt) in all settings except one (Codex with HotpotQA). Finally, we show that even with the much smaller Flan-T5-XXL model, Decomp-Ctxt outperforms all the baselines and can even achieve scores comparable to the Codex-only systems.

\subsection{Additional Results}
\textbf{Post-processing CoT for error correction} 
\acro\ also allows us to create a targeted sub-task handler to focus on the source of error in any system. For example, CoT for arithmetic reasoning often rely on patterns (\texttt{answer is .*}) to extract answers but the CoT does not always fit this pattern. Instead, we can assign the answer extraction to a better sub-task handler (GPT3) and reduce these types of errors. This results in a 17 pt improvement on MultiArith (78 $\to$ 95) and 14 pt improvement on GSM8K (36 $\to$ 50.6) compared to CoT prompting (details in App.~\ref{app:math}).

While \acro\ outperforms the baselines in aggregate, we also see the \textbf{gains of \acro\ are consistent across prompt choices} (see App.~\ref{app:per_prompt}) \textbf{and decomposition schemes} (see App.~\ref{app:scheme}).

\section{Conclusion}
We proposed a new approach, \name, to solve complex tasks using few-shot prompts, by decomposing them into a prompting program built out of simpler sub-tasks. Drawing inspiration from software libraries, our decomposer and shared sub-tasks are designed in a modular fashion: they use their own few-shot prompts, allowing one to independently optimize each prompt, decompose a sub-task further if necessary, or even seamlessly replace it with a symbolic system. We show that \name\ outperforms prior work on four different tasks and generalization settings, establishing it as an effective few-shot paradigm for solving complex tasks.

\section*{Acknowledgements}
We thank members of the Aristo team at the Allen Institute for AI (AI2) for their constructive feedback and the reviewers for their invaluable suggestions. This work was supported in part by the National Science Foundation under grants IIS2007290.

\bibliography{decomp_prompt}
\bibliographystyle{iclr2023_conference}

\clearpage
\appendix

\section{Open Domain QA Details}
\label{app:odqa}

\subsection{Retrieval Corpuses for Open Domain QA}
We use HotpotQA in the fullwiki setting where it comes with the associated Wikipedia corpus for open-domain QA. 2WikiMultihopQA and MuSiQue, however, are originally reading comprehension datasets. Questions in 2WikiMultihopQA and MuSiQue are associated with 10 and 20 paragraphs respectively. To turn these datasets into open-domain QA datasets, we create a corpora for each dataset by combining all the paragraphs in the train, dev and test questions. As a result we get a corpus size of 430,225 paragraphs for 2WikiMultihopQA and 139,416 for MuSiQue.

\subsection{Hyperparameter Tuning for Open Domain QA}

We treat the number of paragraphs to retrieve ($K$) in NoDecomp-Ctxt and Decomp-Ctxt models as a hyperparameter. We select it based on a grid search on a set of values to maximize performance on a held out set of 100 questions for each dataset. For NoDecomp-Ctxt, we search $K \in \{6, 8, 10\}$ for GPT3 models and $K \in {2, 4, 6, 8}$ for Flan-T5-* models. For Decomp-Ctxt, we search $K \in \{2, 4, 6\}$ for GPT3 and Flan-T5-* models. Note that the ranges are different between GPT3 and Flan-T5-* as GPT3 can fit in more number of tokens. The ranges are different for NoDecomp-Ctxt and Decomp-Ctxt as $K$ refers to number of paragraphs retrieved in each round of retrieval, and NoDecomp-Ctxt has only one step of retrieval whereas Decomp-Ctxt usually has multiple retrieval steps. 

\subsection{Additional Results}
\label{app:odqa_results}
\begin{figure}[htbp]
    \centering
    \begin{subfigure}{0.475\textwidth}
    \includegraphics[width=\textwidth]{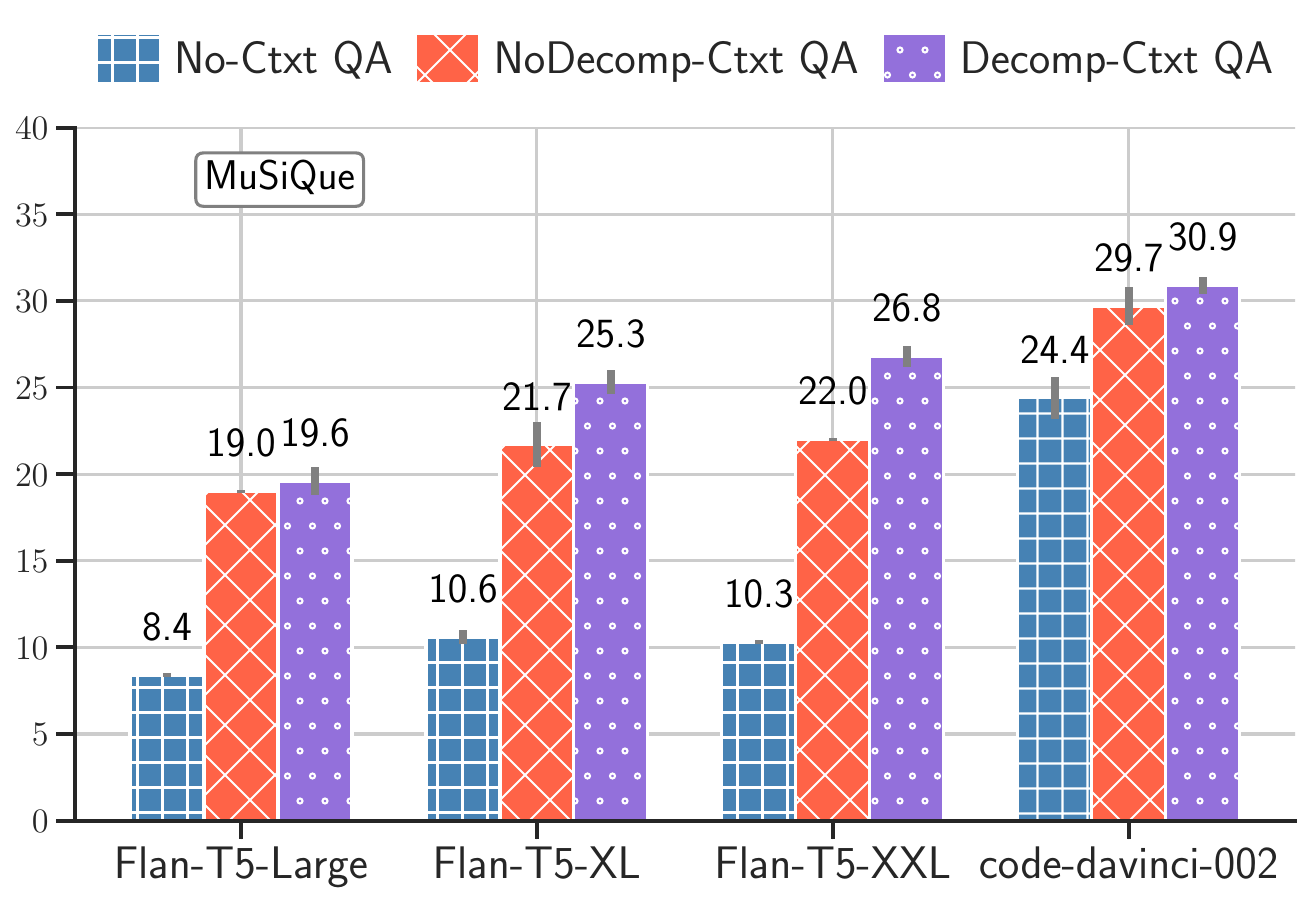}
    \subcaption{CoT QA Prompt}
    \end{subfigure}
    \hfill
    \begin{subfigure}{0.475\textwidth}
    \includegraphics[width=\textwidth]{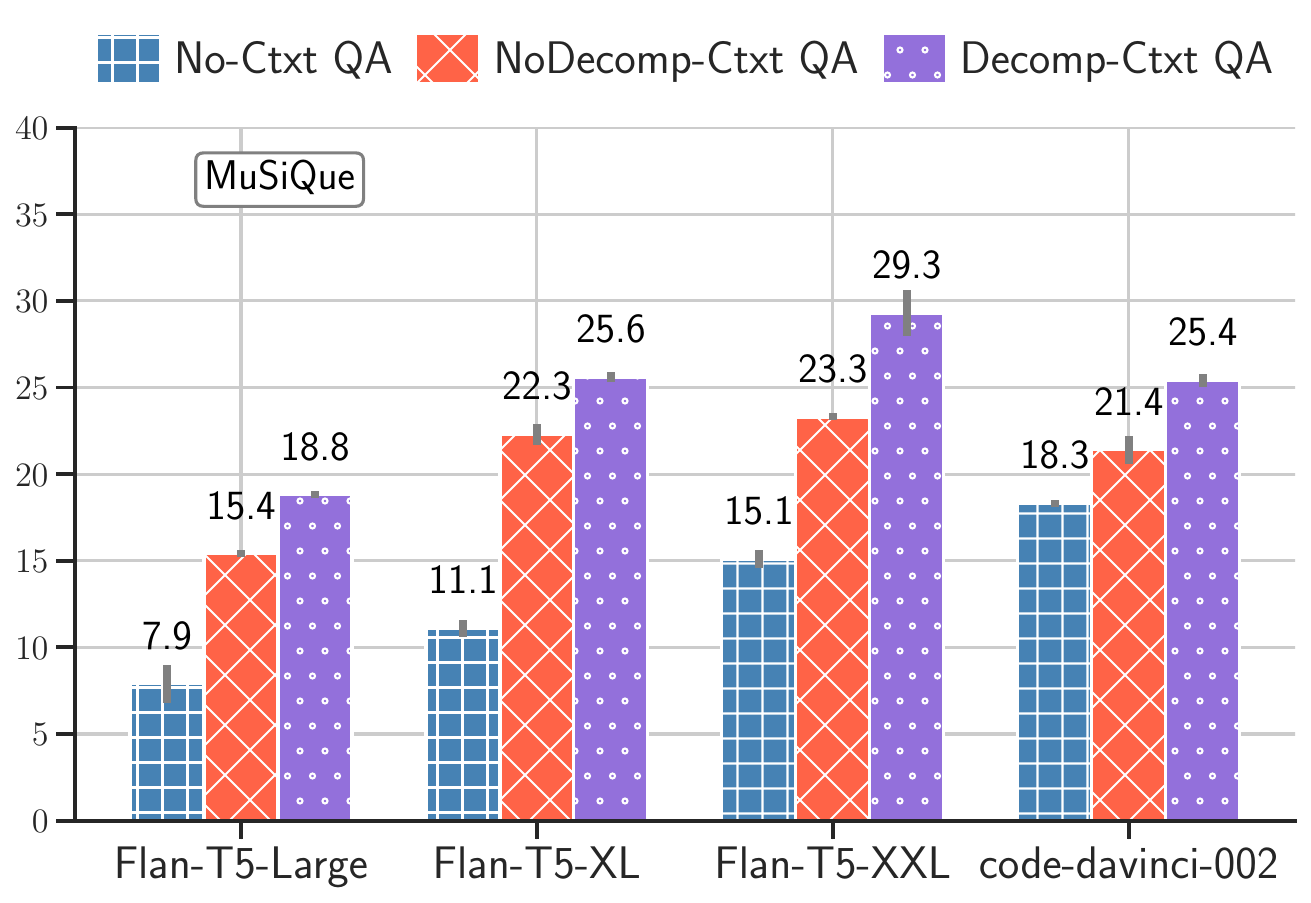}
    \subcaption{Direct QA Prompt}
    \end{subfigure}
    \caption{Results on MuSiQue dataset}
    \label{fig:all_musique_results}
\end{figure}

\begin{figure}[htbp]
    \centering
    \begin{subfigure}{0.475\textwidth}
    \includegraphics[width=\textwidth]{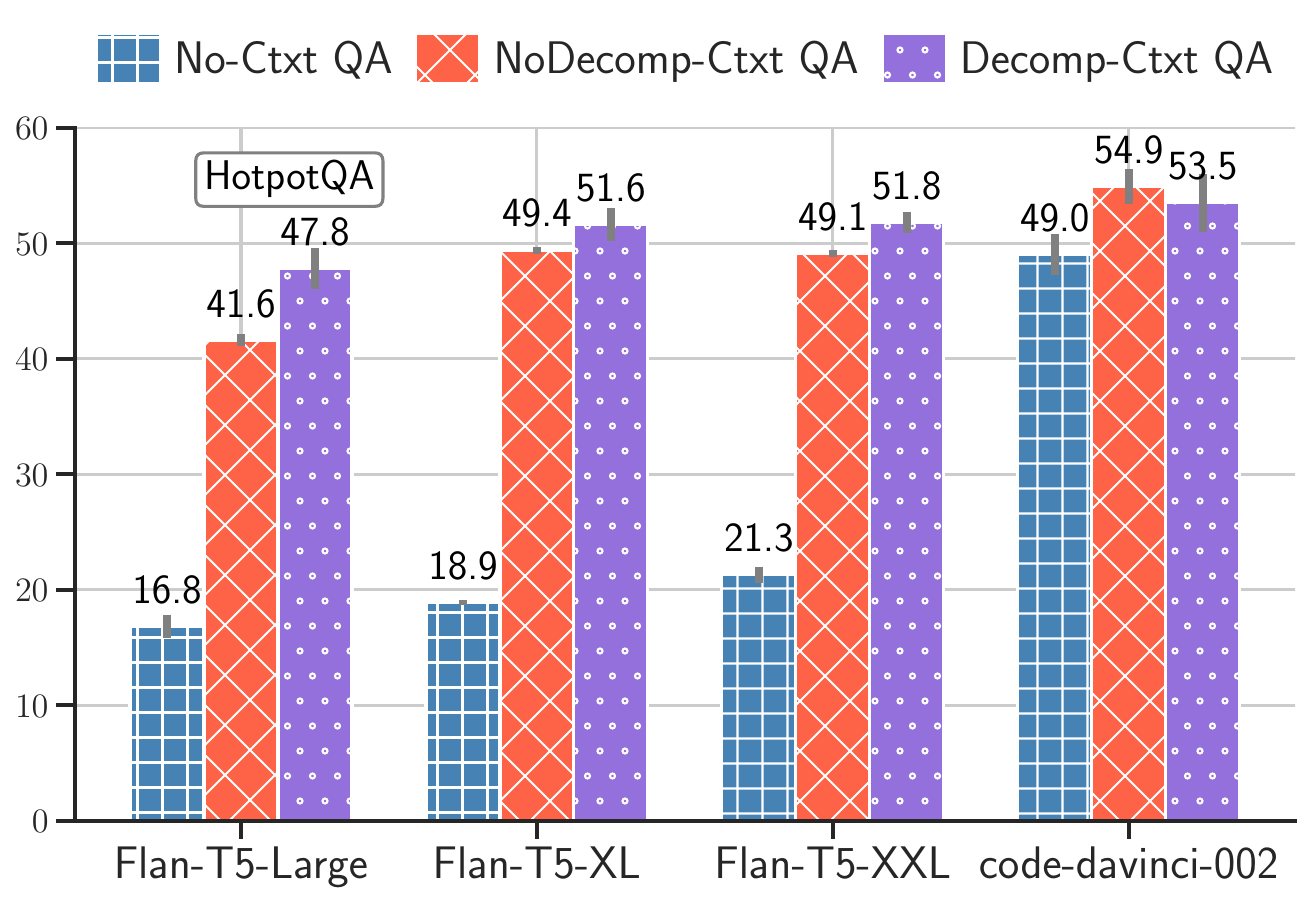}
    \subcaption{CoT QA Prompt}
    \end{subfigure}
    \hfill
    \begin{subfigure}{0.475\textwidth}
    \includegraphics[width=\textwidth]{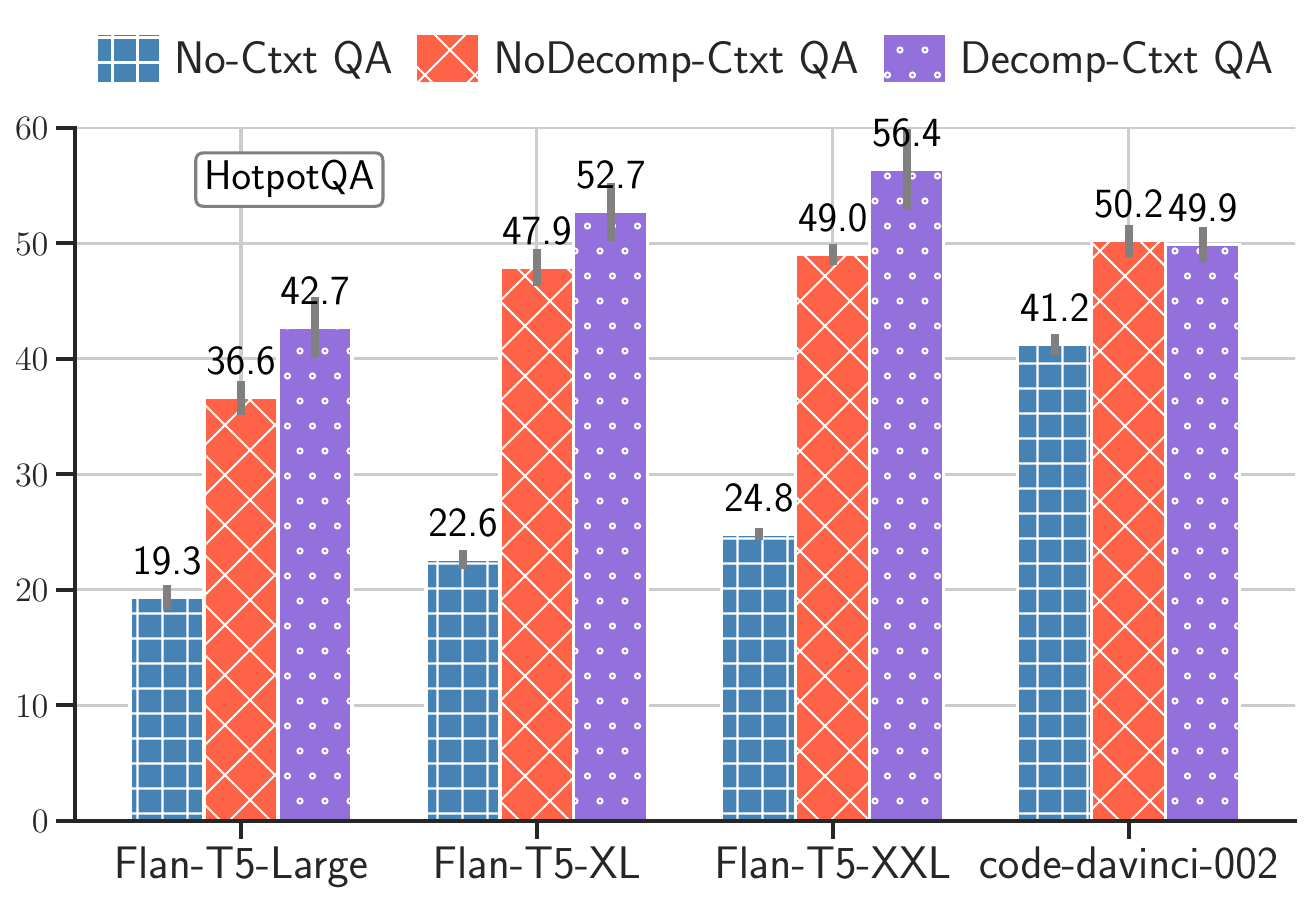}
    \subcaption{Direct QA Prompt}
    \end{subfigure}
    \caption{Results on HotpotQA dataset}
    \label{fig:all_hotpot_results}
\end{figure}

\begin{figure}[htbp]
    \centering
    \begin{subfigure}{0.475\textwidth}
    \includegraphics[width=\textwidth]{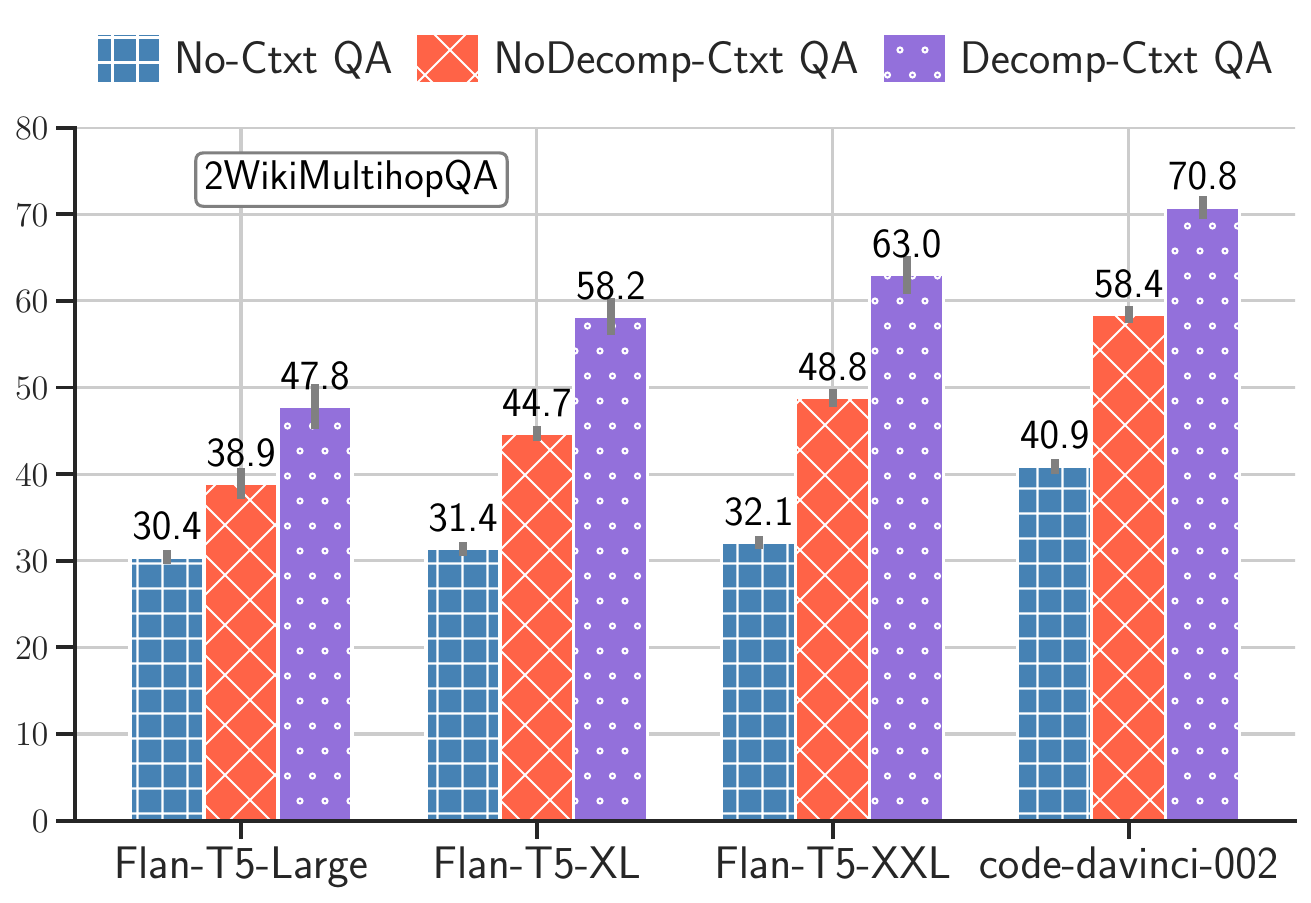}
    \subcaption{CoT QA Prompt}
    \end{subfigure}
    \hfill
    \begin{subfigure}{0.475\textwidth}
    \includegraphics[width=\textwidth]{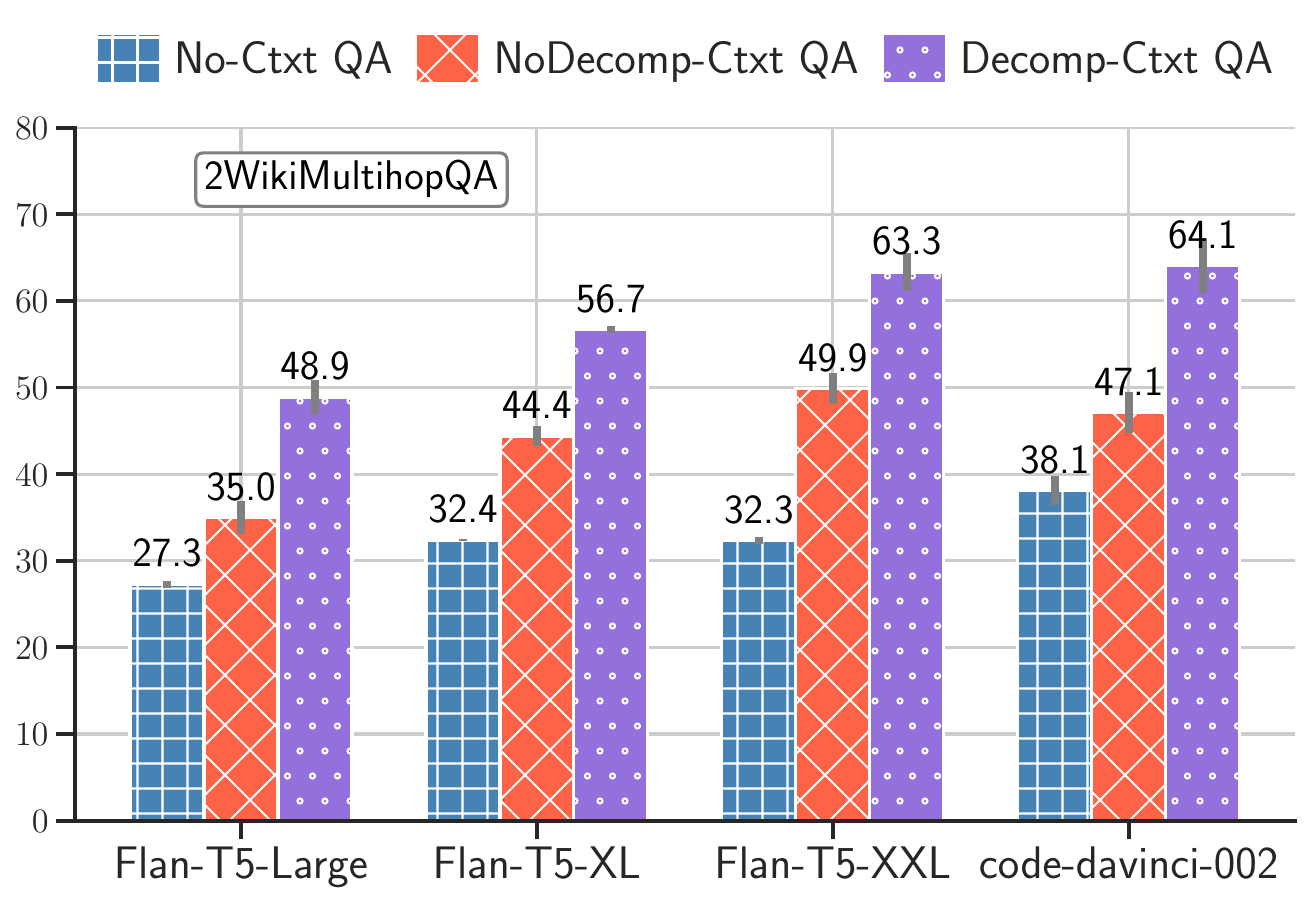}
    \subcaption{Direct QA Prompt}
    \end{subfigure}
    \caption{Results on 2WikiMultihopQA dataset}
    \label{fig:all_2wiki_results}
\end{figure}

\subsubsection{MuSiQue}
We present all the results on the MuSiQue dataset in Fig.~\ref{fig:all_musique_results}. Across all settings, we can see that retrieval helps substantially (large gains over No-Ctxt QA) with further improvements achieved by our DecomP-based Decomp-Ctxt QA model.

\subsubsection{HotpotQA}
We present all the results on the HotpotQA dataset in Fig.~\ref{fig:all_hotpot_results}. On this dataset too, we can see large gains by incorporating retrieval but the gains from using DecomP are mostly seen in the smaller models.

\subsubsection{2WikiMultihopQA}
We present all the results on the 2WikiMultihopQA dataset in Fig.~\ref{fig:all_2wiki_results}. On this dataset, we can see large gains by incorporating retrieval and also observe substantial gains by incorporating DecomP (as compared to NoDecomp-Ctxt).

\section{Math QA}
\label{app:math}
We apply \name\ to two math QA datasets: GSM8K~\cite{cobbe2021gsm8k} and MultiArith~\cite{roy-roth-2015-solving}. For Chain-of-thought, we used the original prompts for math reasoning~\cite{Wei2022ChainOT}. For example:
\textbox{\linewidth}{%
Q: There are 15 trees in the grove. Grove workers will plant trees in the grove today. After they are done, there will be 21 trees. How many trees did the grove workers plant today?
A: There are 15 trees originally. Then there were 21 trees after some more were planted. So there must have been 21 - 15 = 6. The answer is 6.
}

Most CoT systems~\cite{Wei2022ChainOT,Wang2022SelfConsistencyIC} rely on extracting the answer by finding the number following ``answer is''. However, this may not always be accurate. For example, the following CoT would be unanswerable by relying on simple patterns.

\textbox{\linewidth}{%
Parker chews 4 pieces of gum a day. There are 15 pieces of gum in a pack. So he will need 4 * 30 / 15 = 8 packs of gum to last him 30 days.
}

Rather than relying on patterns with limited generalization, we can use a language model to extract the answer more reliably. Specifically, we use \name\ to decompose the task into first identifying the chain-of-thought reasoning and then using a second GPT3-based sub-module to extract the answer from the CoT. We show examples of our prompts here (full prompt in App. ~\ref{app:prompts}):\\

\textbf{Example from the Decomposition Prompt}
\textbox{\linewidth}{%
QC: There are 15 trees in the grove. Grove workers will plant trees in the grove today. After they are done, there will be 21 trees. How many trees did the grove workers plant today?\\
QS: [cot] There are 15 trees in the grove. Grove workers will plant trees in the grove today. After they are done, there will be 21 trees. How many trees did the grove workers plant today?\\
A: There are 15 trees originally. Then there were 21 trees after some more were planted. So there must have been 21 - 15 = 6 trees planted.\\
QS: [gpt\_ans] There are 15 trees originally. Then there were 21 trees after some more were planted. So there must have been 21 - 15 = 6 trees planted.\\
A: 6\\
QS: [EOQ]
}

\textbf{Example from the gpt\_ans prompt}
\textbox{\linewidth}{%
Q: There are 15 trees originally. Then there were 21 trees after some more were planted. So there must have been 21 - 15 = 6 trees planted.\\
A: 6
}

\begin{figure}[htbp]
    \begin{minipage}{0.5\textwidth}
    \includegraphics[width=\textwidth]{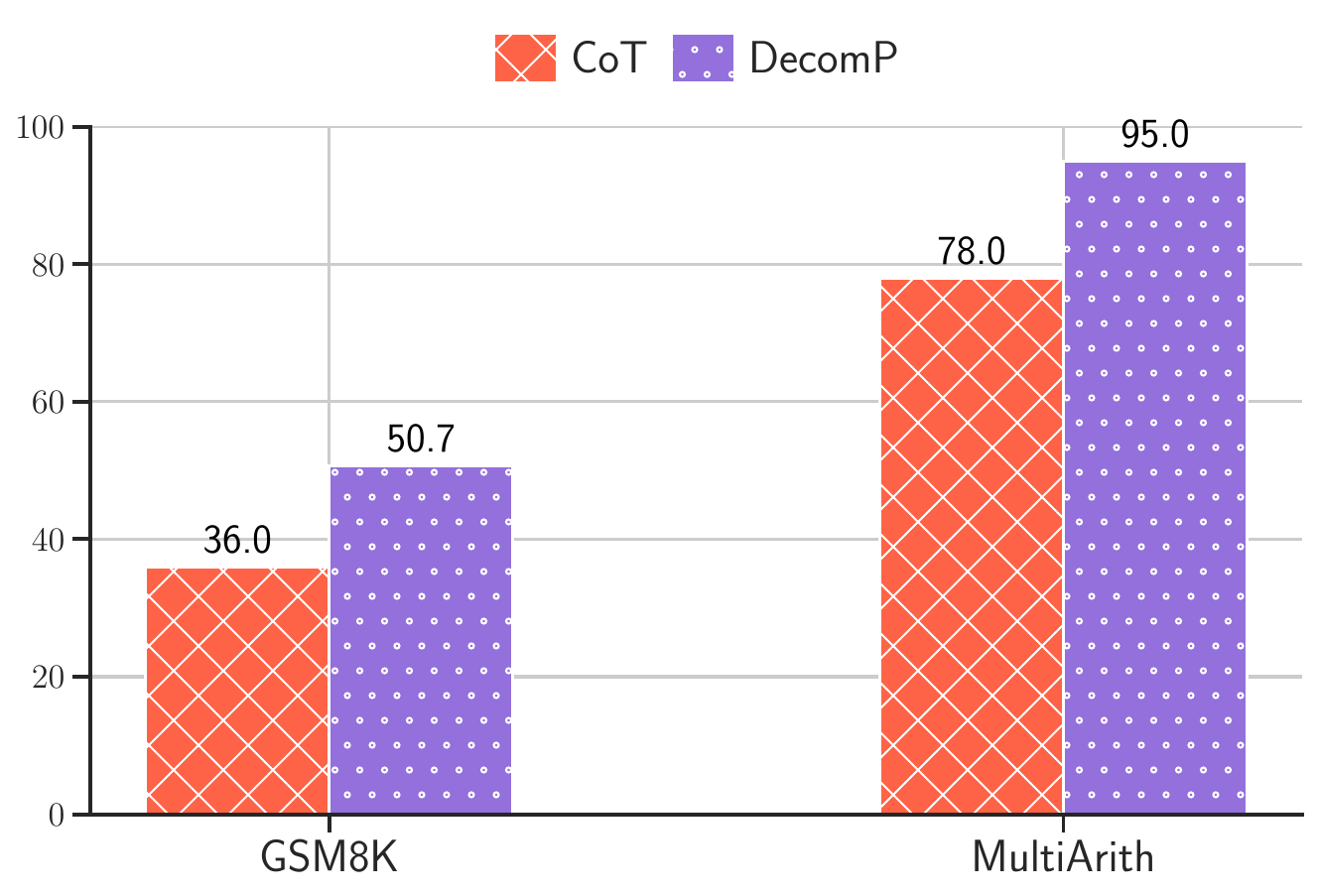}
    \caption{Our simple decomposition results in 14-17 pts on two MathQA datasets: GSM8k and MultiArith.}
    \label{fig:mathqa}
    \end{minipage}
    \hfill
    \begin{minipage}{0.45\textwidth}
    \includegraphics[width=\textwidth]{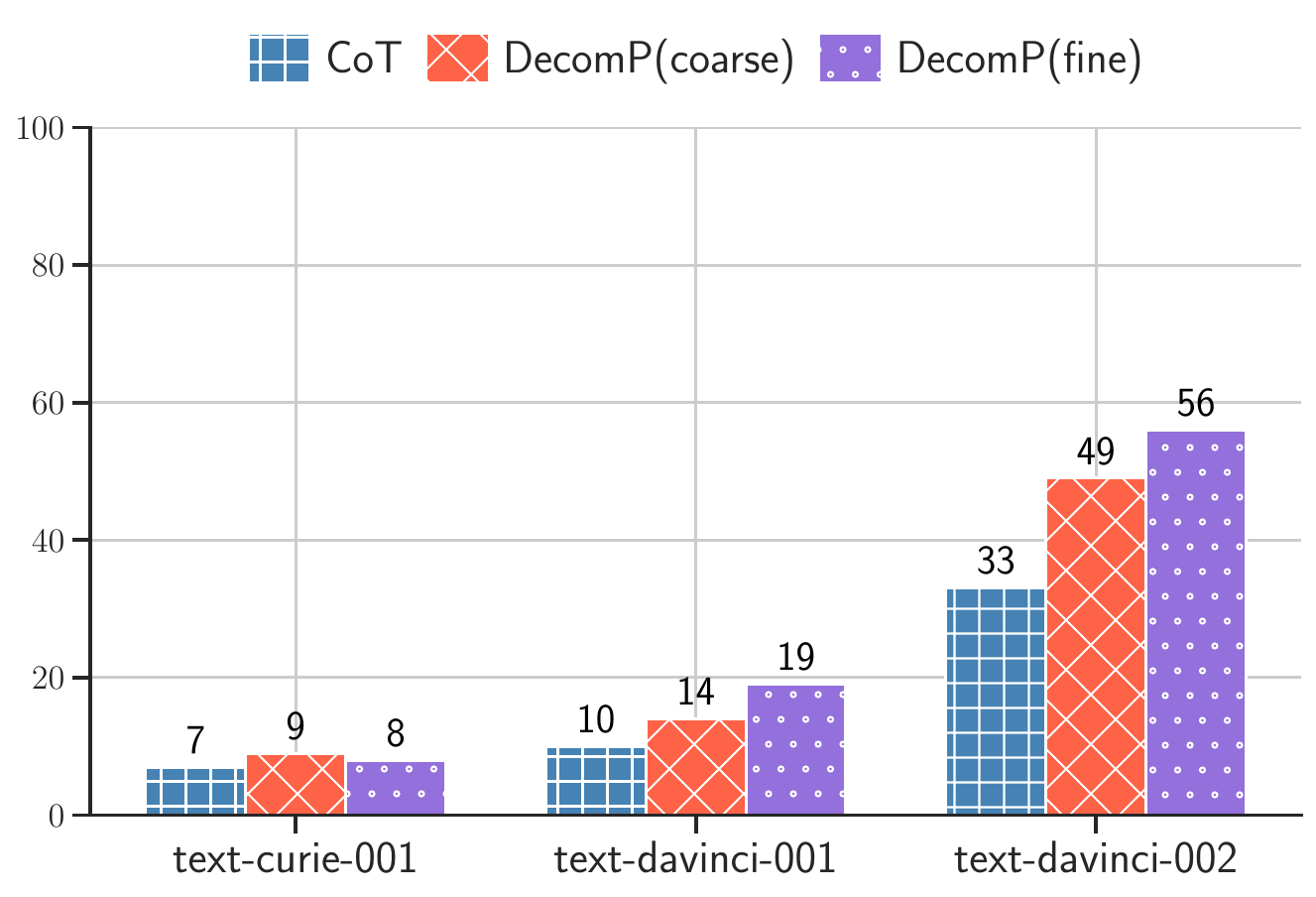}
    \caption{As the models become weaker (davinci-001) and smaller (curie-001), the performance of all the models drop. \acro\ still outperforms CoT till the performance reaches close to zero with curie.}
    \label{fig:scale}
    \end{minipage}
\end{figure}

We present our results in Fig.~\ref{fig:mathqa}. 
On the GSM8K data set\footnote{We randomly sample 300 examples from the test set due to costs with API usage}, we outperform CoT by 14 points. On the MultiArith dataset\footnote{We randomly sample 200 examples from the test set due to costs with API usage}, we achieve a 17 pt improvement compare to CoT. While this is a simple change, it illustrates the possibility of using \acro\ for other complex answer types, e.g. non-extractive answer generation from chain-of-thoughts.

\section{Effect of Scale on CommAQA}
We evaluate text-curie-001, text-davinci-001 and text-davinci-002 on the CommAQA dataset. Since the curie-001 and davinci-001 have a smaller context window size, we further reduced our prompts to fit within their context windows (2048 tokens). As shown in Fig.~\ref{fig:scale}, both CoT and \acro\ are effected by the model size.

\section{Results on all prompts}
\label{app:per_prompt}
\subsection{Per-Prompt Result on Letter Concatenation}
\begin{figure}[htbp]
    \begin{minipage}{.5\textwidth}
    \includegraphics[width=\linewidth]{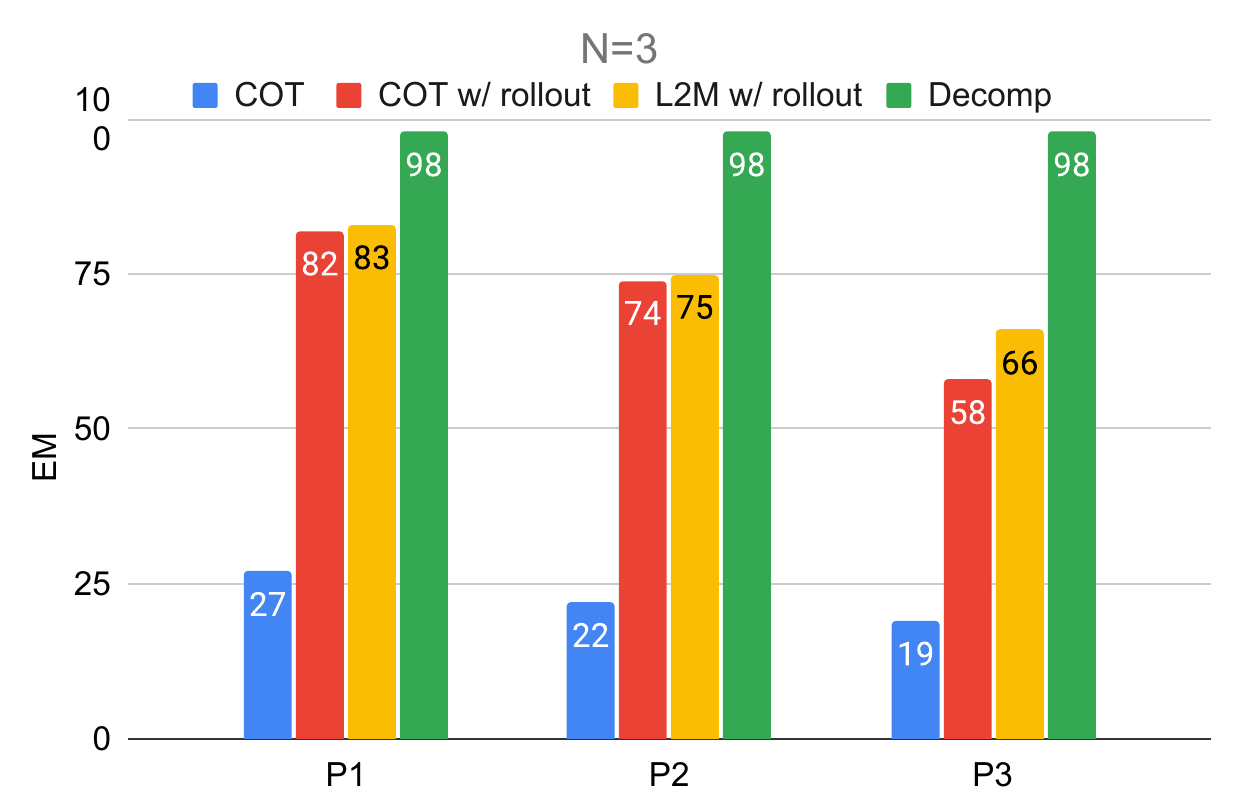}
    \end{minipage}
    \begin{minipage}{.5\textwidth}
    \includegraphics[width=\linewidth]{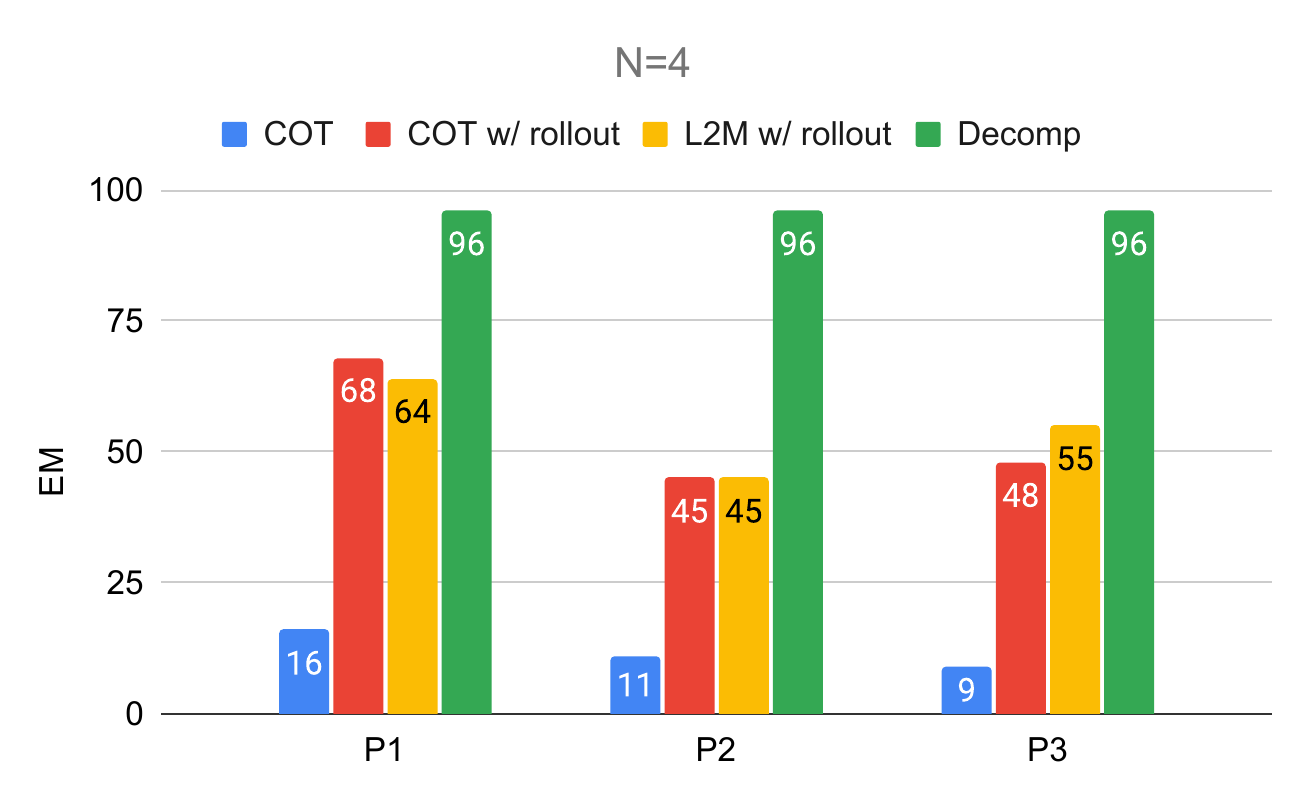}
    \end{minipage}
    \begin{minipage}{.5\textwidth}
    \includegraphics[width=\linewidth]{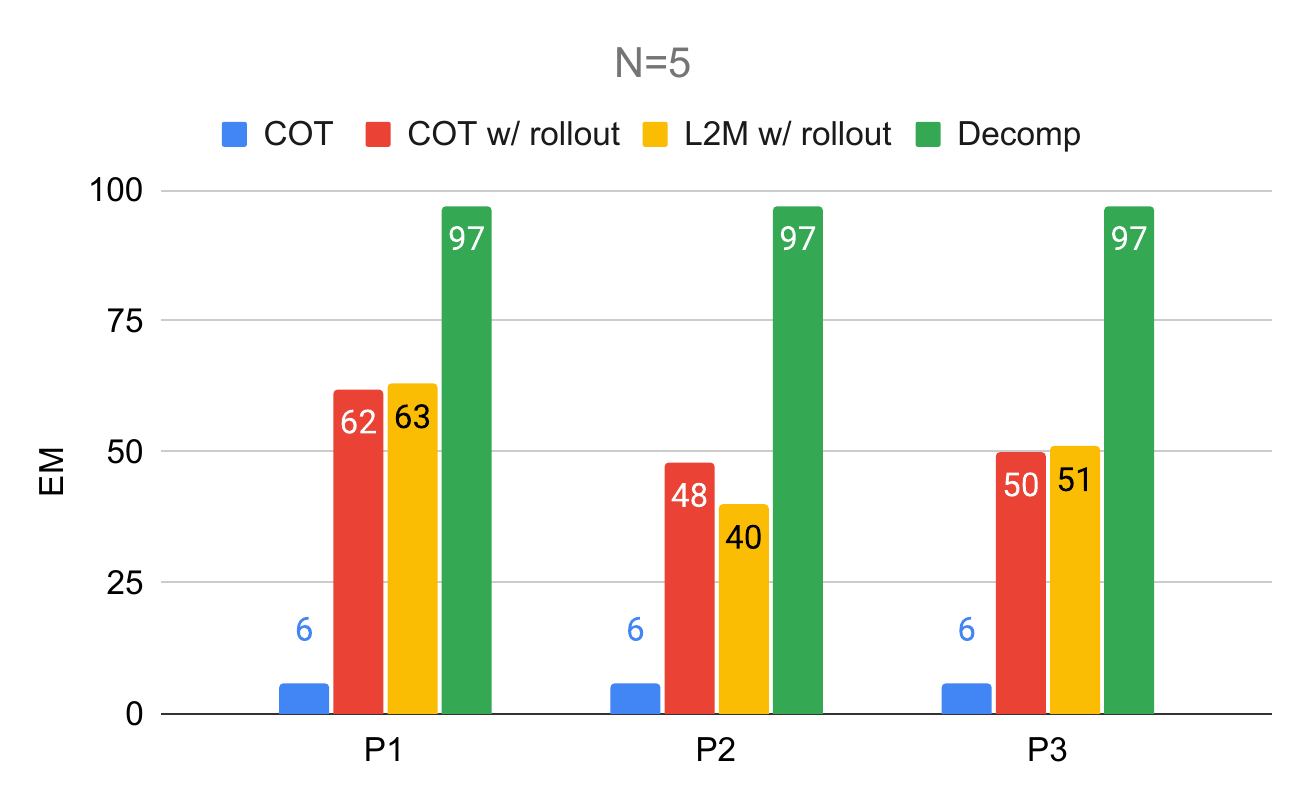}
    \end{minipage}
    \caption{Across all values of N and different prompts (P1, P2 and P3), \acro\ outperform chain-of-thought reasoning and even least-to-most prompting.}
    \label{fig_all_letter_cat}
\end{figure}

We present the results of the letter concatenation task (with space delimiter) for different values of N in Fig.~\ref{fig_all_letter_cat}. Our results are stable across the different prompts (P1, P2 and P3) and always outperform CoT and Least-to-Most prompting. 

\subsection{Per-Prompt Results on CommaQA}
\begin{figure}[htbp]
    \begin{minipage}{.5\textwidth}
    \includegraphics[width=\linewidth]{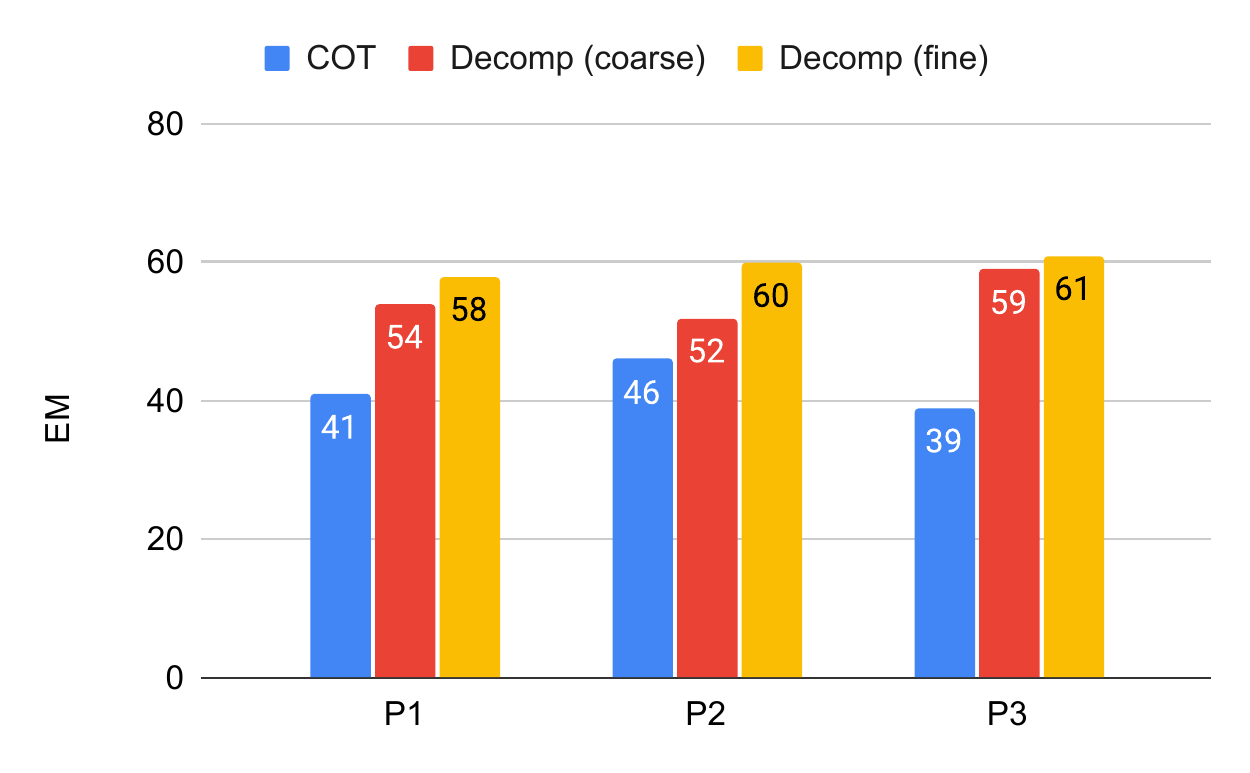}
    \subcaption{Test Set}
    \end{minipage}
    \begin{minipage}{.5\textwidth}
    \includegraphics[width=\linewidth]{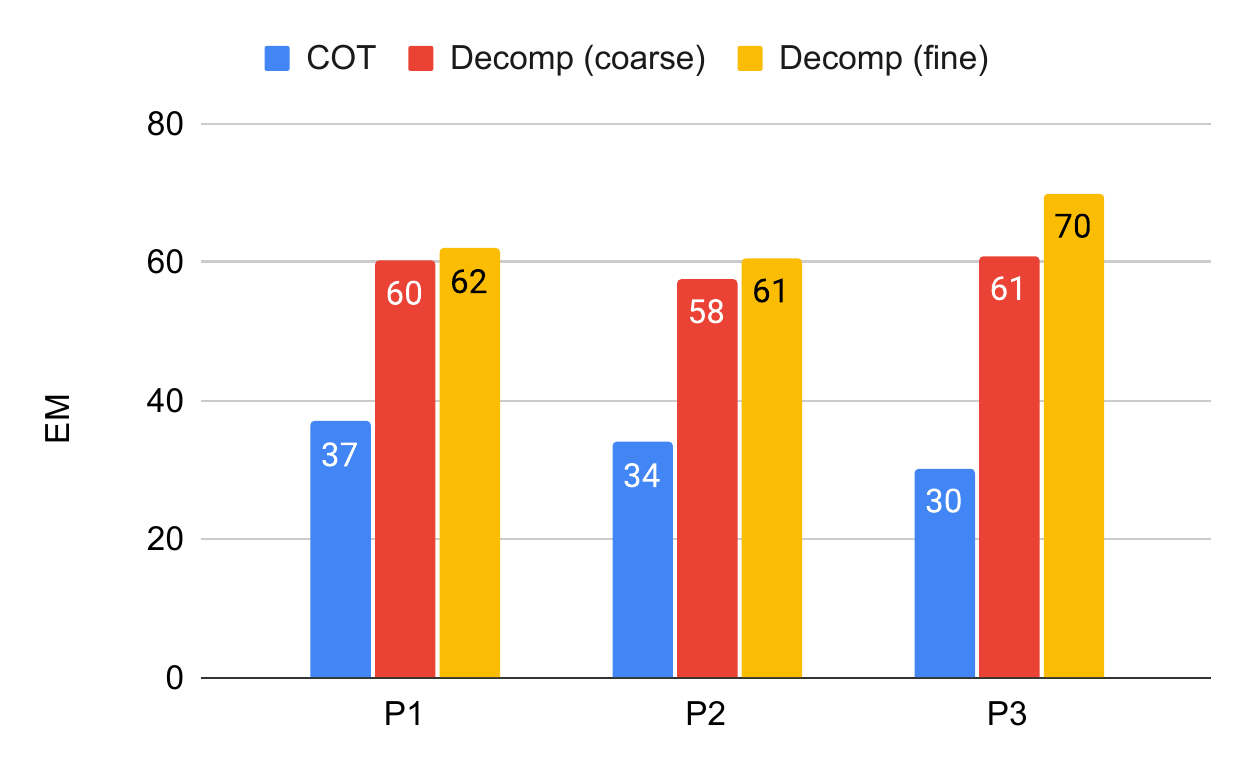}
    \subcaption{Comp. Gen. Set}
    \end{minipage}
    \caption{Results of different prompts on the CommAQA dataset.}
    \label{fig:all_commaqa_results}
\end{figure}

We also present the results of all the prompts on the CommAQA dataset in Fig.~\ref{fig:all_commaqa_results}. Here too, we can observe that \acro\ outperforms CoT on each prompt set.

\section{Effect of Decomposition Scheme}
\label{app:scheme}
 To evaluate the effect of the decomposition scheme, we experiment with two other simple decomposition structures for the letter concatenation and reversal tasks.

\paragraph{Letter Concatenation}
For letter concatenation, we consider an alternate scheme where we use GPT3 to generate each question rather than loop over the answers, e.g.,
\textbox{\linewidth}{%
QC: Take the last letters of the words in "Augusta Ada King" and concatenate them using a space.\\
QS: [split] What are the words in "Augusta Ada King"?\\
A: ["Augusta", "Ada", "King"]\\
QS: [str\_position] What is the last letter in "Augusta"?\\
A: "a"\\
QS: [str\_position] What is the last letter in "Ada"?\\
A: "a"\\
QS: [str\_position] What is the last letter in "King"?\\
A: "g"\\
QS: [merge] Concatenate ["a", "a", "g"] using a space.\\
A: "a a g"\\
QS: [EOQ]
}
By using the decomposer prompt model to generate the sub-questions, we can be more robust to formatting issues in the output answers, e.g., we can expect GPT3 to still generate the appropriate sub-questions even if the first answer is not a valid array. However, the generated sub-questions may not correctly use all the elements of the list (change in order, missed element, repeated elements, etc).

\paragraph{List Reversal}
For list reversal, instead of splitting into halves, we take the tail of the list, reverse it and then concatenate it to the head. i.e. reverse(list) = reverse(list[1:]) + list[0]. This requires more GPT3 calls (O(n)) compared to the original approach of splitting the list into halves (O(log(n))).

\begin{figure}[htbp]
    \begin{minipage}{.475\textwidth}
    \includegraphics[width=\linewidth]{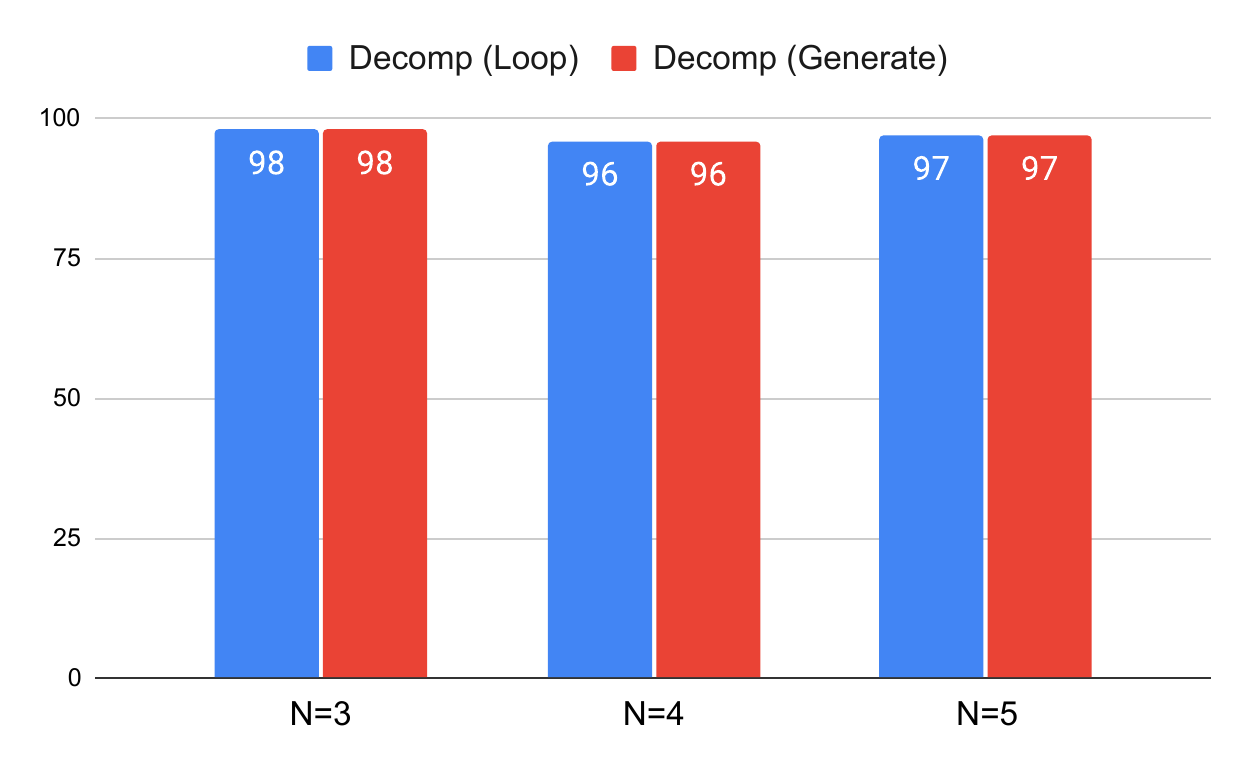}
    \caption{Both decomposition schemes for the letter concatenation task have the same scores.}\label{fig:letter_cat_schema}
    \end{minipage}
    \hfill
    \begin{minipage}{.475\textwidth}
    \includegraphics[width=\linewidth]{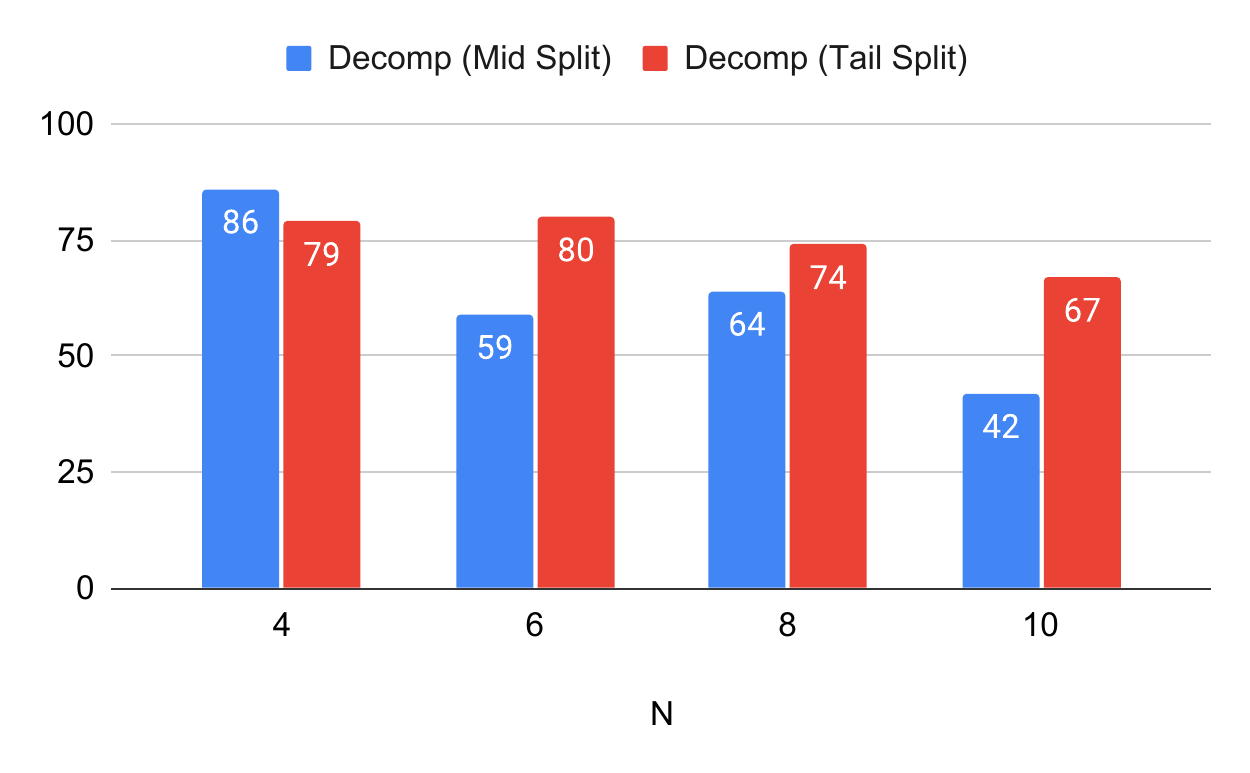}
    \caption{Recursively reversing the tail of  a list is more stable at longer lengths but comes at the cost of more calls to GPT3. }\label{fig:reverse_list_schema}
    \end{minipage}
\end{figure}

In both these cases, we noticed that the performance did not drop as shown in Fig.~\ref{fig:letter_cat_schema} and Fig.~\ref{fig:reverse_list_schema}. On the letter concatenation task, the results were exactly the same.  The new reversal decomposition schema was actually stronger on longer inputs at the cost of more calls to GPT3 (O(ln(n)) using binary splits vs O(n) one element at a time). Both these decomposition schemes are still better than CoT.

\begin{figure}
\centering
\includegraphics[width=0.6\linewidth]{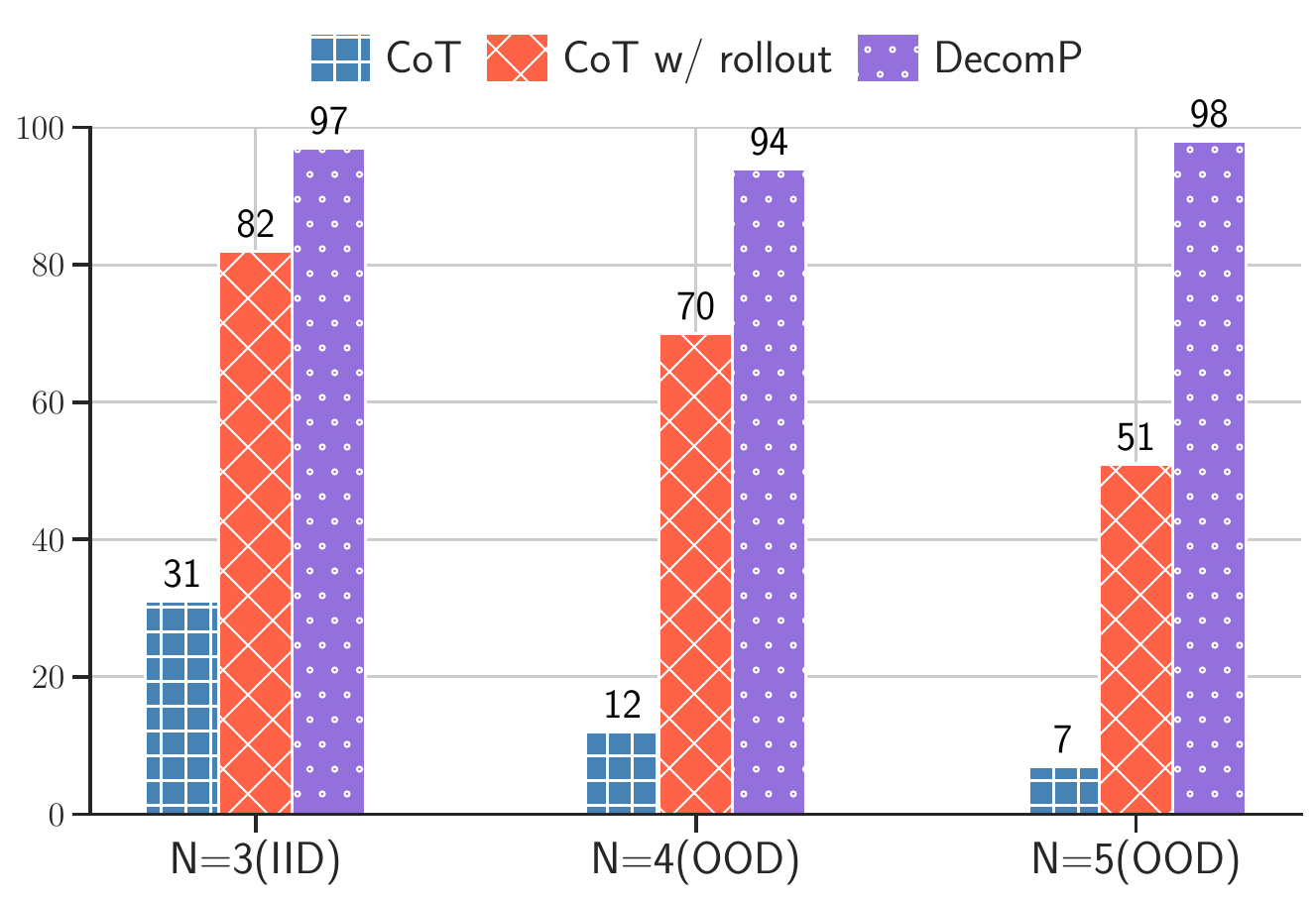}
        \caption{EM Results on the $k^{th}$ letter concatenation task (k=3) using semi-colon as delimiter with different values for N, the number of words in the input. \acro\ always outperforms and generalizes better than \COT.
    }
    \label{fig:letter_cat_semic_results}
\end{figure}

\section{Error Analysis}
\subsection{Letter Concatenation}

\subsubsection{\acro}
We analyzed the errors in \acro\ on the letter concatenation task and only found errors in the sub-task execution. 
\textbox{\linewidth}{%
Q: Take the letters at position 3 of the words in "Nancy Samina Abbas Caudhari Bano" and concatenate them using a space.\\
A: n m b u n\\
Prediction: c m b u n\\
Error: Incorrect letter extraction (Sub-task)\\
{\color{red} \textit{What is at position 3 in "[(N, 1), (a, 2), (n, 3), (c, 4), (y, 5)]"? $\Rightarrow$ "c"}}\\
\\
Q: Take the letters at position 3 of the words in "Orlando Stephen Cho Teixeira Pierre" and concatenate them using a space.\\
A: l e o i e\\
Prediction: leoie\\
Error: Incorrect concatenation (Sub-task)\\
{\color{red} \textit{Concatenate ["l", "e", "o", "i", "e"] using a space. $\Rightarrow$ "leoie"}}
}

\subsubsection{CoT w/ rollout}
We analyzed the errors in CoT on the letter concatenation task and found similar errors during the generation of CoT. But the frequency of these errors was higher than \acro, as it is not possible to effectively teach each sub-task with CoT.
\textbox{\linewidth}{%
Q: Take the letters at position 3 of the words in "Sheila Nicolas Verma Sha Sousa" and concatenate them using a space.\\
A: e c r a u \\
Pred: i c r a u \\
Error: Incorrect letter extraction\\
\textit{...The letters and their positions in "Sheila" are "[(S, 1), (h, 2), (e, 3), (i, 4), (l, 5), (a, 6)]". The letter at position 3 in this sequence is {\color{red}"i"}...}\\
\\
Q:Take the letters at position 3 of the words in "Shobha Kailash Nakamura Peter Benitez" and concatenate them using a space.\\
A: o i k t n\\
Pred: o l k t i\\
Error: Incorrect letter extraction\\
\textit{..."Benitez" are "[(B, 1), (e, 2), (n, 3), (i, 4), (t, 5), (e, 6), (z, 7)]". The letter at position 3 in this sequence is {\color{red}"i"}...}
}

\subsection{CommaQA}
Similarly in CommaQA, the errors are mostly due to sub-task errors, which in this dataset correspond to answering single-hop questions. CoT also makes the same types of errors but they are more frequent since this QA sub-task can not be delegated to a specialized prompt in CoT. Since all errors are of this type, we show only one example here.
\textbox{\linewidth}{%
Q: What awards have movies written by people born in 1933 won?\\
A: ["Hydrallium", "Pompasole"]\\
Pred: ["Pompasole"]\\
Error: Incorrect sub-question answer\\
\textit{Sub-Q:What movies has Haldron written?\\
Sub-A: ["Polytetrafluoromethane", "Skia", "Warpstone"]\\
Pred: ["Skia", "Warpstone"]}
}

\section{Task Prompts}
\label{app:prompts}
We have provided the task prompts for all the datasets for COT and our \name\ approach. 

\paragraph{CoT} Since CoT methods also perform 2-step reasoning: first generate the chain-of-thought and second extract the answer from the CoT, we use the same decomposition-based framework for COT baselines too. For example, consider the following example in our COT prompt:
\textbox{\linewidth}{%
QC: Take the letters at position 1 of the words in ”Alan Mathison Turing” and concatenate them
using a space.\\
QS: [extract] The letter at position 1 of ”Alan” is ”A”. The letter at position 1 of ”Mathison” is ”M”. The
letter at position 1 of ”Turing” is ”T”. Concatenating ”A”, ”M”, ”T” using a space leads to ”A
M T”. So, ”Alan Mathison Turing” outputs ”A M T”.\\
A: ”A M T”\\
QS: [EOQ]\\
}
GPT3 generates the chain-of-thought during the "decomposition" step and a regex-based answer extractor \texttt{extract} (\texttt{'.* outputs "(.*)"\textbackslash.'}) then takes this CoT and generates the answer. In some cases, the module name is skipped in the prompt (the CoT is sent to the extractor by default).

\paragraph{Operators} In this work, we use the same operators as defined by \citeauthor{Khot2022HeyAC}. Their \texttt{select} operator is just the basic operator that replaces references to an answer index with its answer. When not specified, \texttt{select} is assumed to be the default operator. In addition, we consider two operators in our experiments: \texttt{project\_values} and \texttt{project\_values\_flat\_unique}.
\begin{itemize}
    \item \texttt{project\_values}: This operator takes a list answer $\#i = X$ and iterates over it to generate new questions by replacing mentions of $\#i$ i.e. Q = [q.replace(\#i, x) for x $\in$ X]. The answer to each question is simply concatenated to get the final answer i.e. A = [model(q) for q $\in$ Q]. We refer to this as \texttt{foreach} for simplicity in the main text.
    
    \item \texttt{project\_values\_flat\_unique}: This operator performs the same steps as \texttt{project\_values} but then additionally flattens the list and only returns the unique entities in the flattened list. We refer to this as \texttt{foreach\_merge} in the main text for simplicity.
\end{itemize}
\subsection{Letter Concatentation}
We show one of the prompts used for experiments here. The entire set of prompts is provided as supplemetary material.
\colorlet{shadecolor}{gray!10}
\lstset{breaklines=true, columns=fullflexible, backgroundcolor=\color{shadecolor}}

\subsubsection{\name}
\begin{lstlisting}[title=\texttt{decomp}]
QC: Take the last letters of the words in "Augusta Ada King" and concatenate them using a space.
QS: [split] What are the words in "Augusta Ada King"?
A: ["Augusta", "Ada", "King"]
QS: (project_values) [str_position] What is the last letter in "#1"?
A: ["a", "a", "g"]
QS: [merge] Concatenate #2 using a space.
A: "a a g"
QS: [EOQ]

QC: Take the letters at position 1 of the words in "Alan Mathison Turing" and concatenate them using a space.
QS: [split] What are the words in "Alan Mathison Turing"?
A: ["Alan", "Mathison", "Turing"]
QS: (project_values) [str_position] What is the letter at position 1 in "#1"?
A: ["A", "M", "T"]
QS: [merge] Concatenate #2 using a space.
A: "A M T"
QS: [EOQ]

QC: Take the letters at position 4 of the words in "Herbert Alexander Simon" and concatenate them using a space.
QS: [split] What are the words in "Herbert Alexander Simon"?
A: ["Herbert", "Alexander", "Simon"]
QS: (project_values) [str_position] What is the letter at position 4 in "#1"?
A: ["b", "x", "o"]
QS: [merge] Concatenate #2 using a space.
A: "b x o"
QS: [EOQ]
\end{lstlisting}
\begin{lstlisting}[title=\texttt{split}]
Q: What are the words in "Alan Mathison Turing"?
A: ["Alan", "Mathison", "Turing"]

Q: What are the letters in "Alan"?
A: ["A", "l", "a", "n"]

Q: What are the letters and their positions in "Mathison"?
A: "[(M, 1), (a, 2), (t, 3), (h, 4), (i, 5), (s, 6), (o, 7), (n, 8)]"

Q: What are the words and their positions in "Herbert Alexander Simon"?
A: "[(Herbert, 1), (Alexander, 2), (Simon, 3)]"
\end{lstlisting}
\begin{lstlisting}[title=\texttt{str\_position}]
QC: What is the letter at position 1 of the word "Augusta"?
QS: (select) [split] What are the letters and their positions in "Augusta"?
A: "[(A, 1), (u, 2), (g, 3), (u, 4), (s, 5), (t, 6), (a, 7)]"
QS: (select) [arr_position] What is at position 1 in #1?
A: "A"
QS: [EOQ]

QC: What is the last letter of the word "Mathison"?
QS: (select) [split] What are the letters and their positions in "Mathison"?
A: "[(M, 1), (a, 2), (t, 3), (h, 4), (i, 5), (s, 6), (o, 7), (n, 8)]"
QS: (select) [arr_position] What is the last letter in #1?
A: "n"
QS: [EOQ]

QC: What is the word at the position 4 in "Colorless green ideas sleep furiously"?
QS: (select) [split] What are the words and their positions in "Colorless green ideas sleep furiously"?
A: "[(Colorless, 1), (green, 2), (ideas, 3), (sleep, 4), (furiously, 5)]"
QS: (select) [arr_position] What is at the position 4 in #1?
A: "sleep"
QS: [EOQ]
\end{lstlisting}
\begin{lstlisting}[title=\texttt{merge}]
Q: Concatenate ["A", "l", "a", "n"].
A: "Alan"

Q: Concatenate ["b", "x", "o"] using a space.
A: "b x o"

Q: Concatenate ["a", "a", "g"] using a comma.
A: "a,a,g"

Q: Concatenate ["Alan", "Mathison", "Turing"] using a space.
A: "Alan Mathison Turing"

Q: Concatenate ["Allen", "Institute"].
A: "AllenInstitute"
\end{lstlisting}
\begin{lstlisting}[title=\texttt{arr\_position}]
Q: What is at position 4 in "[("Colorless", 1), ("green", 2), ("ideas", 3), ("sleep", 4), ("furiously", 5)]"?
A: "sleep"

Q: What is at position 1 in "[(M, 1), (a, 2), (t, 3), (h, 4), (i, 5), (s, 6), (o, 7), (n, 8)]"?
A: "M"

Q: What is at the last position in "[(A, 1), (u, 2), (g, 3), (u, 4), (s, 5), (t, 6), (a, 7)]"?
A: "a"

Q: What is at position 1 in "[(Herbert, 1), (Alexander, 2), (Simon, 3)]"?
A: "Herbert"

Q: What is at last position in "[(Allen, 1), (Institute, 2), (for, 3), (Artificial, 4), (Intelligence, 5)]"?
A: "Intelligence"

Q: What is at position 4 in "[(A, 1), (l, 2), (e, 3), (x, 4), (a, 5), (n, 6), (d, 7), (e, 8), (r, 9)]"?
A: "x"
\end{lstlisting}
\subsubsection{COT with rollout}
\begin{lstlisting}[title=\texttt{COT w/ rollout}]
QC: Take the last letters of the words in "Augusta Ada King" and concatenate them using a space.
QS: The words in "Augusta Ada King" are "Augusta", "Ada" and "King". The letters and their positions in "Augusta" are "[(A, 1), (u, 2), (g, 3), (u, 4), (s, 5), (t, 6), (a, 7)]". The last letter in this sequence is "a". The letters and their positions in "Ada" are "[(A, 1), (d, 2), (a, 3)]". The last letter in this sequence is "a". The letters and their positions in "King" are "[(K, 1), (i, 2), (n, 3), (g, 4)]". The last letter in this sequence is "g". Concatenating "a", "a", "g" using a space leads to "a a g". So, "Augusta Ada King" outputs "a a g".
A: "a a g"
QS: [EOQ]

QC: Take the letters at position 1 of the words in "Alan Mathison Turing" and concatenate them using a space.
QS: The words in "Alan Mathison Turing" are "Alan", "Mathison",  and "Turing". The letters and their positions in "Alan" are "[(A, 1), (l, 2), (a, 3), (n, 4)]". The letter at position 1 in this sequence is "A". The letters and their positions in "Mathison" are [(M, 1), (a, 2), (t, 3), (h, 4), (i, 5), (s, 6), (o, 7), (n, 8)]". The letter at position 1 in this sequence is "M". The letters and their positions in "Turing" are "[(T, 1), (u, 2), (r, 3), (i, 4), (n, 5), (g, 6)]". The letter at position 1 in this sequence is "T". Concatenating "A", "M", "T" using a space leads to "A M T". So, "Alan Mathison Turing" outputs "A M T".
A: "A M T"
QS: [EOQ]

QC: Take the letters at position 4 of the words in "Herbert Alexander Simon" and concatenate them using a space.
QS: The words in "Herbert Alexander Simon" are "Herbert", "Alexander", and "Simon". The letters and their positions in "Herbert" are "[(H, 1), (e, 2), (r, 3), (b, 4), (e, 5), (r, 6), (t, 7)]". The letter at position 4 in this sequence is "b". The letters and their positions in "Alexander" are "[(A, 1), (l, 2), (e, 3), (x, 4), (a, 5), (n, 6), (d, 7), (e, 8), (r, 9)]". The letter at position 4 in this sequence is "x". The letters and their positions in "Simon" are "[(S, 1), (i, 2), (m, 3), (o, 4), (n, 5)]". The letter at position 4 in this sequence is "o". Concatenating "b", "x", "o" using a space leads to "b x o". So, "Herbert Alexander Simon" outputs "b x o".
A: "b x o"
QS: [EOQ]
\end{lstlisting}
\subsubsection{COT}
\begin{lstlisting}[title=\texttt{COT}]
QC: Take the last letters of the words in "Augusta Ada King" and concatenate them using a space.
QS: The last letter of "Augusta" is "a". The last letter of "Ada" is "a". The last letter of "King" is "g". Concatenating "a", "a", "g" using a space leads to "a a g". So, "Augusta Ada King" outputs "a a g".
A: "a a g"
QS: [EOQ]

QC: Take the letters at position 1 of the words in "Alan Mathison Turing" and concatenate them using a space.
QS: The letter at position 1 of "Alan" is "A". The letter at position 1 of "Mathison" is "M". The letter at position 1 of "Turing" is "T". Concatenating "A", "M", "T" using a space leads to "A M T". So, "Alan Mathison Turing" outputs "A M T".
A: "A M T"
QS: [EOQ]

QC: Take the letters at position 4 of the words in "Herbert Alexander Simon" and concatenate them using a space.
QS: The letter at position 4 of "Herbert" is "b". The letter at position 4 of "Alexander" is "x". The letter at position 4 of "Simon" is "o". Concatenating "b", "x", "o" using a space leads to "b x o". So, "Herbert Alexander Simon" outputs "b x o".
A: "b x o"
QS: [EOQ]
\end{lstlisting}
\subsubsection{Least-to-most w/ rollout}
\begin{lstlisting}[title=\texttt{Least-to-most Decomp}]
QC: Take the last letters of the words in "Augusta Ada King" and concatenate them using a space.
QS: [l2m] Take the last letters of the words in "Augusta Ada" and concatenate them using a space.
A: The words in "Augusta Ada" are "Augusta" and "Ada". The letters and their positions in "Augusta" are "[(A, 1), (u, 2), (g, 3), (u, 4), (s, 5), (t, 6), (a, 7)]". The last letter in this sequence is "a". The letters and their positions in "Ada" are "[(A, 1), (d, 2), (a, 3)]". The last letter in this sequence is "a". Concatenating "a", "a" using a space leads to "a a". So, "Augusta Ada" outputs "a a".
QS: [l2m] Take the last letters of the words in "Augusta Ada King" and concatenate them using a space.
A: "Augusta Ada" outputs "a a". The letters and their positions in "King" are "[(K, 1), (i, 2), (n, 3), (g, 4)]". The last letter in this sequence is "g". Concatenating "a a", "g" using a space leads to "a a g". So, "Augusta Ada King" outputs "a a g".
QS: [ans_ext] So, "Augusta Ada King" outputs "a a g".
A: "a a g"
QS: [EOQ]

QC: Take the letters at position 1 of the words in "Alan Mathison Turing" and concatenate them using a space.
QS: [l2m] Take the letters at position 1 of the words in "Alan Mathison" and concatenate them using a space.
A: The words in "Alan Mathison" are "Alan" and "Mathison". The letters and their positions in "Alan" are "[(A, 1), (l, 2), (a, 3), (n, 4)]". The letter at position 1 in this sequence is "A". The letters and their positions in "Mathison" are [(M, 1), (a, 2), (t, 3), (h, 4), (i, 5), (s, 6), (o, 7), (n, 8)]". The letter at position 1 in this sequence is "M". Concatenating "A", "M" using a space leads to "A M". So, "Alan Mathison" outputs "A M".
QS: [l2m] Take the letters at position 1 of the words in "Alan Mathison Turing" and concatenate them using a space.
A: "Alan Mathison" outputs "A M". The letters and their positions in "Turing" are "[(T, 1), (u, 2), (r, 3), (i, 4), (n, 5), (g, 6)]". The letter at position 1 in this sequence is "T". Concatenating "A M", "T" using a space leads to "A M T". So, "Alan Mathison Turing" outputs "A M T".
QS: [ans_ext] So, "Alan Mathison Turing" outputs "A M T".
A: "A M T"
QS: [EOQ]

QC: Take the letters at position 4 of the words in "Herbert Alexander Simon" and concatenate them using a space.
QS: [l2m] Take the letters at position 4 of the words in "Herbert Alexander" and concatenate them using a space.
A: The words in "Herbert Alexander" are "Herbert" and "Alexander". The letters and their positions in "Herbert" are "[(H, 1), (e, 2), (r, 3), (b, 4), (e, 5), (r, 6), (t, 7)]". The letter at position 4 in this sequence is "b". The letters and their positions in "Alexander" are "[(A, 1), (l, 2), (e, 3), (x, 4), (a, 5), (n, 6), (d, 7), (e, 8), (r, 9)]". The letter at position 4 in this sequence is "x". Concatenating "b", "x" using a space leads to "b x". So, "Herbert Alexander" outputs "b x".
QS: [l2m] Take the letters at position 4 of the words in "Herbert Alexander Simon" and concatenate them using a space.
A: "Herbert Alexander" outputs "b x". The letters and their positions in "Simon" are "[(S, 1), (i, 2), (m, 3), (o, 4), (n, 5)]". The letter at position 4 in this sequence is "o". Concatenating "b x", "o" using a space leads to "b x o". So, "Herbert Alexander Simon" outputs "b x o".
QS: [ans_ext] So, "Herbert Alexander Simon" outputs "b x o".
A: "b x o"
QS: [EOQ]
\end{lstlisting}
\begin{lstlisting}[title=\texttt{Least-to-most COT(l2m)}]
Q: Take the last letters of the words in "Augusta Ada" and concatenate them using a space.
A: The words in "Augusta Ada" are "Augusta" and "Ada". The letters and their positions in "Augusta" are "[(A, 1), (u, 2), (g, 3), (u, 4), (s, 5), (t, 6), (a, 7)]". The last letter in this sequence is "a". The letters and their positions in "Ada" are "[(A, 1), (d, 2), (a, 3)]". The last letter in this sequence is "a". Concatenating "a", "a" using a space leads to "a a". So, "Augusta Ada" outputs "a a".

Q: Take the last letters of the words in "Augusta Ada King" and concatenate them using a space.
A: "Augusta Ada" outputs "a a". The letters and their positions in "King" are "[(K, 1), (i, 2), (n, 3), (g, 4)]". The last letter in this sequence is "g". Concatenating "a a", "g" using a space leads to "a a g". So, "Augusta Ada King" outputs "a a g".

Q: Take the letters at position 1 of the words in "Alan Mathison" and concatenate them using a space.
A: The words in "Alan Mathison" are "Alan" and "Mathison". The letters and their positions in "Alan" are "[(A, 1), (l, 2), (a, 3), (n, 4)]". The letter at position 1 in this sequence is "A". The letters and their positions in "Mathison" are [(M, 1), (a, 2), (t, 3), (h, 4), (i, 5), (s, 6), (o, 7), (n, 8)]". The letter at position 1 in this sequence is "M". Concatenating "A", "M" using a space leads to "A M". So, "Alan Mathison" outputs "A M".

Q: Take the letters at position 1 of the words in "Alan Mathison Turing" and concatenate them using a space.
A: "Alan Mathison" outputs "A M". The letters and their positions in "Turing" are "[(T, 1), (u, 2), (r, 3), (i, 4), (n, 5), (g, 6)]". The letter at position 1 in this sequence is "T". Concatenating "A M", "T" using a space leads to "A M T". So, "Alan Mathison Turing" outputs "A M T".

Q: Take the letters at position 4 of the words in "Herbert Alexander" and concatenate them using a space.
A: The words in "Herbert Alexander" are "Herbert" and "Alexander". The letters and their positions in "Herbert" are "[(H, 1), (e, 2), (r, 3), (b, 4), (e, 5), (r, 6), (t, 7)]". The letter at position 4 in this sequence is "b". The letters and their positions in "Alexander" are "[(A, 1), (l, 2), (e, 3), (x, 4), (a, 5), (n, 6), (d, 7), (e, 8), (r, 9)]". The letter at position 4 in this sequence is "x". Concatenating "b", "x" using a space leads to "b x". So, "Herbert Alexander" outputs "b x".

Q: Take the letters at position 4 of the words in "Herbert Alexander Simon" and concatenate them using a space.
A: "Herbert Alexander" outputs "b x". The letters and their positions in "Simon" are "[(S, 1), (i, 2), (m, 3), (o, 4), (n, 5)]". The letter at position 4 in this sequence is "o". Concatenating "b x", "o" using a space leads to "b x o". So, "Herbert Alexander Simon" outputs "b x o".
\end{lstlisting}

\subsubsection{Alt \acro\ schema (Generate Each Sub-Question)}
\begin{lstlisting}[title=\texttt{decomp}]
QC: Take the last letters of the words in "Augusta Ada King" and concatenate them using a space.
QS: [split] What are the words in "Augusta Ada King"?
A: ["Augusta", "Ada", "King"]
QS: [str_position] What is the last letter in "Augusta"?
A: "a"
QS: [str_position] What is the last letter in "Ada"?
A: "a"
QS: [str_position] What is the last letter in "King"?
A: "g"
QS: [merge] Concatenate ["a", "a", "g"] using a space.
A: "a a g"
QS: [EOQ]

QC: Take the letters at position 1 of the words in "Alan Mathison Turing" and concatenate them using a space.
QS: [split] What are the words in "Alan Mathison Turing"?
A: ["Alan", "Mathison", "Turing"]
QS: [str_position] What is the letter at position 1 in "Alan"?
A: "A"
QS: [str_position] What is the letter at position 1 in "Mathison"?
A: "M"
QS: [str_position] What is the letter at position 1 in "Turing"?
A: "T"
QS: [merge] Concatenate ["A", "M", "T"] using a space.
A: "A M T"
QS: [EOQ]

QC: Take the letters at position 4 of the words in "Herbert Alexander Simon" and concatenate them using a space.
QS: [split] What are the words in "Herbert Alexander Simon"?
A: ["Herbert", "Alexander", "Simon"]
QS: [str_position] What is the letter at position 4 in "Herbert"?
A: "b"
QS: [str_position] What is the letter at position 4 in "Alexander"?
A: "x"
QS: [str_position] What is the letter at position 4 in "Simon"?
A: "o"
QS: [merge] Concatenate ["b", "x", "o"] using a space.
A: "b x o"
QS: [EOQ]
\end{lstlisting}

\subsection{Sequence Reversal}

\subsubsection{Split Reversal}
The prompts in this section implement Algorithm~\ref{alg:split}.

\begin{algorithm}
  \caption{A recursive reversal strategy that splits the sequence in half, reverses each half, and concatenates them. Runs in $O(\log n)$ calls to the LM where $n$ is the number of items in the sequence.}\label{alg:split}
  	\begin{algorithmic}
		\Procedure{SplitReverse}{$x$}\Comment{The reversal of $x$}
		\If {$|x|<4$}\Comment{The base case}
      \State \Return{$x_{|x|},\ldots,x_1$}\Comment{The reversed sequence}
    \Else\Comment{The inductive case}
      \State $n\gets |x|/2$\Comment{Half the length of $x$}
      \State $\ell\gets x_1,\ldots,x_n$\Comment{The first half of $x$}
      \State $\ell^R\gets \Call{SplitReverse}{\ell}$\Comment{The reversed first half}
      \State $r\gets x_{n+1},\ldots,x_{|x|}$\Comment{The second half of $x$}
      \State{$r^R\gets \Call{SplitReverse}{r}$}\Comment{The reversed second half}
      \State \Return{$r^R_1,\ldots,r^R_n,\ell^R_{n+1},\ldots,\ell^R_{|x|}$}\Comment{The concatenated reversed halves}
    \EndIf
    \EndProcedure
	\end{algorithmic}

\end{algorithm}

\begin{lstlisting}[title=\texttt{reverse}]
QC: Reverse the sequence "driving license, button, packet, identity card, shoe".
QS: [extract] The sequence is "1. driving license, 2. button, 3. packet, 4. identity card, 5. shoe". The sequence is 5 items long, which is more than the minimum length of 4, so we split it. Half of 5 is 5 / 2 = 2.5. Dropping the decimal, we get that the first half will be 2 items long, ending in "2. button". The first half (2 items) is "1. driving license, 2. button".
A: "1. driving license, 2. button"
QS: [extract] The first half of the sequence ends with "2. button", so the second half starts after "2. button" with "3. packet". The full sequence is 5 items long, and the first half is 2 items long, so the second half will be 5 - 2 = 3 items long. The second half of the sequence (3 items) is "3. packet, 4. identity card, 5. shoe".
A: "3. packet, 4. identity card, 5. shoe"
QS: [remove_numbers] Remove the numbers from #1.
A: "driving license, button"
QS: [remove_numbers] Remove the numbers from #2.
A: "packet, identity card, shoe"
QS: [reverse] Reverse the sequence #3. 
A: "button, driving license"
QS: [reverse] Reverse the sequence #4.
A: "shoe, identity card, packet"
QS: [join] #6 #5
A: "shoe, identity card, packet, button, driving license"
QS: [EOQ]

QC: Reverse the sequence "laptop, photo, clip".
QS: [extract] The sequence is "1. laptop, 2. photo, 3. clip". The sequence is 3 items long, which is less than the minimum length of 4, so we don't need to split it. All we need to do is reverse "laptop, photo, clip".
A: "laptop, photo, clip"
QS: [cot] Reverse the sequence #1.
A: "clip, photo, laptop"
QS: [EOQ]

QC: Reverse the sequence "newspaper, glasses, laptop, bottle".
QS: [extract] The sequence is "1. newspaper, 2. glasses, 3. laptop, 4. bottle". The sequence is 4 items long, which is equal to the minimum length of 4, so we split it. Half of 4 is 4 / 2 = 2.0. Dropping the decimal, we get that the first half will be 2 items long. The first half (2 items) of the sequence is "1. newspaper, 2. glasses".
A: "1. newspaper, 2. glasses"
QS: [extract] The first half of the sequence ends with "2. glasses", so the second half starts after "2. glasses" with "3. laptop". The full sequence is 4 items long and the first half is 2 items long, so the second half will be 4 - 2 = 2 items long, ending in "2. glasses". The second half of the sequence (2 items) is "3. laptop, 4. bottle".
A: "3. laptop, 4. bottle"
QS: [remove_numbers] Remove the numbers from #1.
A: "newspaper, glasses"
QS: [remove_numbers] Remove the numbers from #2.
A: "laptop, bottle"
QS: [reverse] Reverse the sequence #3. 
A: "glasses, newspaper"
QS: [reverse] Reverse the sequence #4.
A: "bottle, laptop"
QS: [join] #6 #5
A: "bottle, laptop, glasses, newspaper"
QS: [EOQ]
\end{lstlisting}
\begin{lstlisting}[title=\texttt{remove\_numbers}]
Q: Remove the numbers from "4. bottle, 3. laptop, 2. glasses, 1. newspaper".
A: "bottle, laptop, glasses, newspaper"

Q: Remove the numbers from "1. identity card, 2. packet, 3. button".
A: "identity card, packet, button"

Q: Remove the numbers from "1. player, 2. passport, 3. umbrella, 4. radio".
A:  "player, passport, umbrella, radio"
\end{lstlisting}
\begin{lstlisting}[title=\texttt{join}]
Q: "bottle, laptop" "glasses, newspaper".
A: "bottle, laptop, glasses, newspaper"

Q: "identity card, packet, button" "magazine, notebook, glasses".
A: "identity card, packet, button, magazine, notebook, glasses"

Q: "passport, umbrella, radio, mobile phone, photo" "player".
A:  "passport, umbrella, radio, mobile phone, photo, player"

Q: "mirror, case" "toothbrush, alarm clock".
A: "mirror, case, toothbrush, alarm clock"

Q: "light bulb, clip, umbrella" "driving licence, watch".
A: "light bulb, clip, umbrella, driving licence, watch"
\end{lstlisting}
\begin{lstlisting}[title=\texttt{cot}]
QC: Reverse the sequence "newspaper, glasses, laptop, bottle".
QS: First is newspaper. Second is glasses. Third is laptop. Fourth is bottle. Now to reverse, change the order to: Fourth is bottle. Third is laptop. Second is glasses. First is newspaper. So the answer is "bottle, laptop, glasses, newspaper".
A: "bottle, laptop, glasses, newspaper"

QC: Reverse the sequence "laptop, photo, clip".
QS: First is laptop. Second is photo. Third is clip. Now to reverse, change the order to: Third is clip. Second is photo. First is laptop. So the answer is "clip, photo, laptop".
A: "clip, photo, laptop"

QC: Reverse the sequence "driving license, button, packet, identity card, pineapple".
QS: First is driving license. Second is button. Third is packet. Fourth is identity card. Fifth is pineapple. Now to reverse, change the order to: Fifth is pineapple. Fourth is identity card. Third is packet. Second is button. First is driving license. So the answer is "pineapple, identity card, packet, button, driving license".
A: "pineapple, identity card, packet, button, driving license"
\end{lstlisting}
\begin{lstlisting}[title=\texttt{unrolled\_decomp}]
QC: Reverse the sequence "driving license, button, packet, identity card, shoe".
QS: The sequence is "1. driving license, 2. button, 3. packet, 4. identity card, 5. shoe". The sequence is 5 items long, which is more than the minimum length of 4, so we split it. Half of 5 is 5 / 2 = 2.5. Dropping the decimal, we get that the first half will be 2 items long, ending in "2. button". The first half (2 items) is "1. driving license, 2. button". The first half of the sequence ends with "2. button", so the second half starts after "2. button" with "3. packet". The full sequence is 5 items long, and the first half is 2 items long, so the second half will be 5 - 2 = 3 items long. The second half of the sequence (3 items) is "3. packet, 4. identity card, 5. shoe". Removing the numbers from "1. driving license, 2. button", we get "driving license, button". Removing the numbers from "3. packet, 4. identity card, 5. shoe", we get "packet, identity card, shoe". Reversing the sequence "driving license, button", First is driving license. Second is button. Now to reverse, change the order to: Second is button. First is driving license. So the answer is "button, driving license". Reversing the sequence "packet, identity card, shoe", First is packet. Second is identity card. Third is shoe. Now to reverse, change the order to: Third is shoe. Second is identity card. First is packet. So the answer is "shoe, identity card, packet". Joining "shoe, identity card, packet" and "button, driving license", the answer is "shoe, identity card, packet, button, driving license".
A: "shoe, identity card, packet, button, driving license"
QS: [EOQ]

QC: Reverse the sequence "laptop, photo, clip".
QS: The sequence is "1. laptop, 2. photo, 3. clip". The sequence is 3 items long, which is less than the minimum length of 4, so we don't need to split it. All we need to do is reverse "laptop, photo, clip". First is laptop. Second is photo. Third is clip. Now to reverse, change the order to: Third is clip. Second is photo. First is laptop. So the answer is "clip, photo, laptop".
A: "clip, photo, laptop"
QS: [EOQ]

QC: Reverse the sequence "newspaper, glasses, laptop, bottle".
QS: The sequence is "1. newspaper, 2. glasses, 3. laptop, 4. bottle". The sequence is 4 items long, which is equal to the minimum length of 4, so we split it. Half of 4 is 4 / 2 = 2.0. Dropping the decimal, we get that the first half will be 2 items long, ending in "2. glasses". The first half (2 items) is "1. newspaper, 2. glasses". The first half of the sequence ends with "2. glasses", so the second half starts after "2. glasses" with "3. laptop". The full sequence is 4 items long, and the first half is 2 items long, so the second half will be 4 - 2 = 2 items long. The second half of the sequence (2 items) is "3. laptop, 4. bottle". Removing the numbers from "1. newspaper, 2. glasses", we get "newspaper, glasses". Removing the numbers from "3. laptop, 4. bottle", we get "laptop, bottle". Reversing the sequence "newspaper, glasses", First is newspaper. Second is glasses. Now to reverse, change the order to: Second is glasses. First is newspaper. So the answer is "glasses, newspaper". Reversing the sequence "laptop, bottle", First is laptop. Second is bottle. Now to reverse, change the order to: Second is bottle. First is laptop. So the answer is "bottle, laptop". Joining "bottle, laptop" and "glasses, newspaper", the answer is "bottle, laptop, glasses, newspaper".
A: "bottle, laptop, glasses, newspaper"
QS: [EOQ]
\end{lstlisting}
\begin{lstlisting}[title=\texttt{reverse (tail)}]
QC: Reverse the sequence "driving license, button, packet, identity card, shoe".
Q1: [extract] The sequence is "1. driving license, 2. button, 3. packet, 4. identity card, 5. shoe". The sequence is 5 items long, which is more than the minimum length of 4, so we split it. We take the first element in the sequence which is "1. driving license".
#1: "1. driving license"
Q2: [extract] The full sequence is 5 items long, so the remaining sequence will be 5 - 1 = 4 items long. The tail of the sequence with 4 items is "2. button, 3. packet, 4. identity card, 5. shoe".
#2: "2. button, 3. packet, 4. identity card, 5. shoe"
Q3: [remove_numbers] Remove the numbers from #1.
#3: "driving license"
Q4: [remove_numbers] Remove the numbers from #2.
#4: "button, packet, identity card, shoe"
Q5: [reverse] Reverse the sequence #4.
#5: "shoe, identity card, packet, button"
Q6: [join] #5 #3
#6: "shoe, identity card, packet, button, driving license"
Q7: [EOQ]

QC: Reverse the sequence "laptop, photo, clip".
Q1: [extract] The sequence is "1. laptop, 2. photo, 3. clip". The sequence is 3 items long, which is less than the minimum length of 4, so we don't need to split it. All we need to do is reverse "laptop, photo, clip".
#1: "laptop, photo, clip"
Q2: [cot] Reverse the sequence #1.
#2: "clip, photo, laptop"
Q3: [EOQ]

QC: Reverse the sequence "newspaper, glasses, laptop, bottle".
Q1: [extract] The sequence is "1. newspaper, 2. glasses, 3. laptop, 4. bottle". The sequence is 4 items long, which is more than the minimum length of 4, so we split it. We take the first element in the sequence which is "1. newspaper".
#1: "1. newspaper"
Q2: [extract] The full sequence is 4 items long, so the remaining sequence will be 4 - 1 = 3 items long. The tail of the sequence with 3 items is "2. glasses, 3. laptop, 4. bottle".
#2: "2. glasses, 3. laptop, 4. bottle"
Q3: [remove_numbers] Remove the numbers from #1.
#3: "newspaper"
Q4: [remove_numbers] Remove the numbers from #2.
#4: "glasses, laptop, bottle"
Q5: [reverse] Reverse the sequence #4.
#5: "bottle, laptop, glasses"
Q6: [join] #5 #3
#6: "bottle, laptop, glasses, newspaper"
Q7: [EOQ]
\end{lstlisting}

\subsection{Long-Document QA}
We show one of the prompts used for CommaQA experiments here. The entire set of prompts is provided as supplemetary material.
\subsubsection{\name: (coarse)}
\lstset{literate={è}{{\`e}}1}
\begin{lstlisting}[title=\texttt{decomp}]
What awards have movies produced by people born in 1910 won?
QS: (select) [qa] Who were born in the year 1910?
A: ["Teeplemole", "Muntaril"]
QS: (project_values_flat_unique) [qa] For which movies was #1 the producer?
A: ["Featsaw", "Zalate", "Premercy"]
QS: (project_values_flat_unique) [qa] Which awards were given to #2?
A: ["Zorgion", "Chowwurst", "Hallowcock"]
QS: [EOQ]

QC: What movies have people from the country Stridery acted in?
QS: (select) [qa] Who is from the country Stridery?
A: ["Gastrat"]
QS: (project_values_flat_unique) [qa] Which movies has #1 been an actor in?
A: ["Partnershipmaker", "Nilitude", "Warpstone"]
QS: [EOQ]

QC: What awards have the actors of the Erowid winning movies received?
QS: (select) [qa] Which movies were given the Erowid award?
A: ["Dewbar", "Caudacite"]
QS: (project_values_flat_unique) [qa] Who are the actors in the movie #1?
A: ["Wetherality", "Lougerière", "Gigabut"]
QS: (project_values_flat_unique) [qa] Which awards were given to #2?
A: ["Aniconder", "Trifogation"]
QS: [EOQ]

QC: What awards did the movies directed by the Modiparity winners receive?
QS: (select) [qa] Who has been awarded the Modiparity award?
A: ["Bioperatology"]
QS: (project_values_flat_unique) [qa] Which movies has #1 directed?
A: ["Pestok", "Vitrilateral"]
QS: (project_values_flat_unique) [qa] Which awards were given to #2?
A: ["Gutney", "Antipositive"]
QS: [EOQ]

QC: What awards have movies written by people born in 1935 won?
QS: (select) [qa] Who were born in the year 1935?
A: ["Sclerocybin", "Zayage"]
QS: (project_values_flat_unique) [qa] What movies has #1 written?
A: ["Noenometer", "Tayenne", "Pneumodendron"]
QS: (project_values_flat_unique) [qa] Which awards were given to #2?
A: ["Brownbeard", "Goosehead", "Handt"]
QS: [EOQ]

QC: What movies have the directors from Legault directed?
QS: (select) [qa] Who is from the country Legault?
A: ["Metatoun", "Sapien"]
QS: (project_values_flat_unique) [qa] What movies has #1 been the director of?
A: ["Coacheship", "Misapportionment"]
QS: [EOQ]
\end{lstlisting}
\begin{lstlisting}[title=\texttt{qa}]
movie: Premercy ; directed by: Muntaril. movie: Skirtsicine ; director: Teeplemole. movie: Featsaw ; directed by: Monsterscar. movie: Zalate ; director: Monsterscar. movie: Zalate ; awarded: Hallowcock. movie: Featsaw ; awarded: Zorgion. movie: Premercy ; award: Chowwurst. movie: Skirtsicine ; award: Hallowcock. award: Goatfly ; winner: Teeplemole. person: Monsterscar ; award: Glodome. person: Muntaril ; award: Goatfly. movie: Featsaw ; release year: 1973. movie: Zalate ; release year: 1964. movie: Skirtsicine ; release year: 1973. movie: Premercy ; year: 1961. Teeplemole was an actor in the movie Skirtsicine. Muntaril was an actor in the movie Skirtsicine. Monsterscar was an actor in the movie Premercy. Muntaril was an actor in the movie Featsaw. Teeplemole was an actor in the movie Zalate. Muntaril was born in the year 1910. Teeplemole was born in 1910. Monsterscar was born in 1942. Teeplemole is from the country of Piperfish. Monsterscar is from the country of Piperfish. Muntaril is from the country of Clony. Muntaril produced the movie Skirtsicine with others. Monsterscar was one of the producers of the movie Featsaw. Monsterscar produced the movie Premercy with others. Monsterscar produced the movie Zalate with others. Teeplemole was one of the producers of the movie Featsaw. Teeplemole produced the movie Zalate with others. Muntaril produced the movie Premercy with others. Monsterscar wrote for the movie Premercy. Muntaril was one of the writers for the movie Zalate. Muntaril wrote for the movie Featsaw. Teeplemole wrote for the movie Featsaw. Monsterscar was one of the writers for the movie Zalate. Teeplemole was one of the writers for the movie Skirtsicine.
Q: For which movies was Teeplemole the producer?
A: ["Featsaw", "Zalate"]

Q: Which awards were given to Featsaw?
A: ["Zorgion"]

movie: Misgendery ; directed by: Wetherality. movie: Dewbar ; director: Gigabut. movie: Caudacite ; director: Lougerière. movie: Tayenne ; directed by: Lougerière. movie: Misgendery ; awarded: Microsouenesis. movie: Dewbar ; awarded: Erowid. movie: Tayenne ; awarded: Cockspit. movie: Caudacite ; award: Erowid. award: Aniconder ; winner: Wetherality. award: Aniconder ; winner: Lougerière. person: Gigabut ; award: Trifogation. movie: Dewbar ; release year: 1991. movie: Tayenne ; year: 2013. movie: Caudacite ; release year: 2008. movie: Misgendery ; year: 1991. Wetherality was an actor in the movie Dewbar. Gigabut was an actor in the movie Tayenne. Lougerière was an actor in the movie Tayenne. Lougerière acted in the movie Caudacite. Lougerière acted in the movie Misgendery. Gigabut was an actor in the movie Caudacite. Wetherality was an actor in the movie Misgendery. Wetherality was born in the year 1917. Lougerière was born in 1926. Gigabut was born in the year 1917. Gigabut grew up in the nation of Triclops. Lougerière is from the country of Tatkin. Wetherality grew up in the nation of Tatkin. Lougerière produced the movie Dewbar with others. Gigabut produced the movie Tayenne with others. Gigabut produced the movie Dewbar with others. Lougerière was one of the producers of the movie Misgendery. Wetherality was one of the producers of the movie Caudacite. Gigabut was one of the producers of the movie Caudacite. Wetherality produced the movie Misgendery with others. Wetherality produced the movie Tayenne with others. Wetherality wrote for the movie Tayenne. Gigabut wrote for the movie Misgendery. Lougerière was one of the writers for the movie Caudacite. Wetherality wrote for the movie Misgendery. Gigabut wrote for the movie Tayenne. Gigabut wrote for the movie Dewbar. Lougerière wrote for the movie Dewbar. Wetherality wrote for the movie Caudacite.
Q: Who are the actors in the movie Caudacite?
A: ["Lougerière, "Gigabut"]

Q: Which movies were given the Erowid award?
A: ["Dewbar", "Caudacite"]

movie: Pastillobox ; directed by: Firmline. movie: Clenestration ; directed by: Carblock. movie: Pestok ; directed by: Bioperatology. movie: Vitrilateral ; director: Bioperatology. movie: Vitrilateral ; award: Antipositive. movie: Clenestration ; awarded: Handt. movie: Pastillobox ; awarded: Handt. movie: Pestok ; awarded: Gutney. movie: Pestok ; writer: Firmline. movie: Clenestration ; written by: Carblock. movie: Pastillobox ; written by: Bioperatology. movie: Pestok ; writer: Bioperatology. movie: Clenestration ; written by: Firmline. movie: Vitrilateral ; writer: Bioperatology. movie: Pastillobox ; writer: Carblock. movie: Vitrilateral ; written by: Carblock. movie: Pestok ; release year: 1986. movie: Clenestration ; year: 1986. movie: Vitrilateral ; year: 1999. movie: Pastillobox ; release year: 1984. Carblock was an actor in the movie Pastillobox. Firmline was an actor in the movie Vitrilateral. Bioperatology was an actor in the movie Clenestration. Firmline acted in the movie Pastillobox. Carblock was an actor in the movie Clenestration. Bioperatology was an actor in the movie Pestok. Firmline was born in the year 1904. Bioperatology was born in the year 1935. Carblock was born in 1935. Carblock grew up in the nation of Knoppock. Firmline grew up in the nation of Tatkin. Bioperatology grew up in the nation of Tatkin. Bioperatology won the Modiparity award. Halfbill was awarded to Firmline. Halfbill was awarded to Carblock. Bioperatology was one of the producers of the movie Pestok. Bioperatology produced the movie Vitrilateral with others. Firmline produced the movie Pastillobox with others. Firmline produced the movie Clenestration with others. Carblock was one of the producers of the movie Pastillobox. Carblock produced the movie Vitrilateral with others. Carblock produced the movie Clenestration with others. Firmline was one of the producers of the movie Pestok.
Q: Who has been awarded the Modiparity award?
A: ["Bioperatology"]

Q: Which movies has Bioperatology directed?
A: ["Pestok", "Vitrilateral"]

movie: Nohit ; director: Mimicocycle. movie: Noenometer ; director: Mimicocycle. movie: Tayenne ; directed by: Zayage. movie: Pneumodendron ; director: Sclerocybin. movie: Tayenne ; awarded: Goosehead. movie: Nohit ; awarded: Handt. movie: Pneumodendron ; award: Handt. movie: Noenometer ; awarded: Brownbeard. movie: Nohit ; writer: Mimicocycle. movie: Noenometer ; written by: Sclerocybin. movie: Tayenne ; writer: Sclerocybin. movie: Pneumodendron ; written by: Zayage. movie: Tayenne ; writer: Zayage. movie: Pneumodendron ; written by: Mimicocycle. movie: Noenometer ; release year: 1991. movie: Tayenne ; year: 2013. movie: Nohit ; year: 2005. movie: Pneumodendron ; year: 2005. Mimicocycle was an actor in the movie Tayenne. Zayage acted in the movie Pneumodendron. Zayage was an actor in the movie Nohit. Sclerocybin was an actor in the movie Nohit. Sclerocybin was an actor in the movie Tayenne. Mimicocycle was an actor in the movie Noenometer. Zayage was born in 1935. Sclerocybin was born in 1935. Mimicocycle was born in 1930. Mimicocycle is from the country of Calderita. Sclerocybin grew up in the nation of Calderita. Zayage is from the country of Obility. Quinion was awarded to Zayage. Fannyxist was awarded to Sclerocybin. Fannyxist was awarded to Mimicocycle. Mimicocycle produced the movie Nohit with others. Zayage was one of the producers of the movie Nohit. Sclerocybin was one of the producers of the movie Tayenne. Sclerocybin produced the movie Pneumodendron with others. Zayage produced the movie Pneumodendron with others. Mimicocycle was one of the producers of the movie Tayenne. Sclerocybin was one of the producers of the movie Noenometer. Zayage produced the movie Noenometer with others
Q: What movies has Sclerocybin written?
A: ["Noenometer", "Tayenne"]

Q: Who were born in the year 1935?
A: ["Sclerocybin", "Zayage"]
\end{lstlisting}

\subsubsection{\name: (fine)}
\begin{lstlisting}[title=\texttt{decomp}]
What awards have movies produced by people born in 1910 won?
QS: (select) [simp_qa] Who were born in the year 1910?
A: ["Teeplemole", "Muntaril"]
QS: (project_values_flat_unique) [pos_qa] For which movies was #1 the producer?
A: ["Featsaw", "Zalate", "Premercy"]
QS: (project_values_flat_unique) [aw_qa] Which awards were given to #2?
A: ["Zorgion", "Chowwurst", "Hallowcock"]
QS: [EOQ]

QC: What movies have people from the country Stridery acted in?
QS: (select) [simp_qa] Who is from the country Stridery?
A: ["Gastrat"]
QS: (project_values_flat_unique) [pos_qa] Which movies has #1 been an actor in?
A: ["Partnershipmaker", "Nilitude", "Warpstone"]
QS: [EOQ]

QC: What awards have the actors of the Erowid winning movies received?
QS: (select) [aw_qa] Which movies were given the Erowid award?
A: ["Dewbar", "Caudacite"]
QS: (project_values_flat_unique) [pos_qa] Who are the actors in the movie #1?
A: ["Wetherality", "Lougerière", "Gigabut"]
QS: (project_values_flat_unique) [aw_qa] Which awards were given to #2?
A: ["Aniconder", "Trifogation"]
QS: [EOQ]

QC: What awards did the movies directed by the Modiparity winners receive?
QS: (select) [aw_qa] Who has been awarded the Modiparity award?
A: ["Bioperatology"]
QS: (project_values_flat_unique) [pos_qa] Which movies has #1 directed?
A: ["Pestok", "Vitrilateral"]
QS: (project_values_flat_unique) [aw_qa] Which awards were given to #2?
A: ["Gutney", "Antipositive"]
QS: [EOQ]

QC: What awards have movies written by people born in 1935 won?
QS: (select) [simp_qa] Who were born in the year 1935?
A: ["Sclerocybin", "Zayage"]
QS: (project_values_flat_unique) [pos_qa] What movies has #1 written?
A: ["Noenometer", "Tayenne", "Pneumodendron"]
QS: (project_values_flat_unique) [aw_qa] Which awards were given to #2?
A: ["Brownbeard", "Goosehead", "Handt"]
QS: [EOQ]

QC: What movies have the directors from Legault directed?
QS: (select) [simp_qa] Who is from the country Legault?
A: ["Metatoun", "Sapien"]
QS: (project_values_flat_unique) [pos_qa] What movies has #1 been the director of?
A: ["Coacheship", "Misapportionment"]
QS: [EOQ]
\end{lstlisting}
\begin{lstlisting}[title=\texttt{aw\_qa}]
movie: Premercy ; directed by: Muntaril. movie: Skirtsicine ; director: Teeplemole. movie: Featsaw ; directed by: Monsterscar. movie: Zalate ; director: Monsterscar. movie: Zalate ; awarded: Hallowcock. movie: Featsaw ; awarded: Zorgion. movie: Premercy ; award: Chowwurst. movie: Skirtsicine ; award: Hallowcock. award: Goatfly ; winner: Teeplemole. person: Monsterscar ; award: Glodome. person: Muntaril ; award: Goatfly. movie: Featsaw ; release year: 1973. movie: Zalate ; release year: 1964. movie: Skirtsicine ; release year: 1973. movie: Premercy ; year: 1961. Teeplemole was an actor in the movie Skirtsicine. Muntaril was an actor in the movie Skirtsicine. Monsterscar was an actor in the movie Premercy. Muntaril was an actor in the movie Featsaw. Teeplemole was an actor in the movie Zalate. Muntaril was born in the year 1910. Teeplemole was born in 1910. Monsterscar was born in 1942. Teeplemole is from the country of Piperfish. Monsterscar is from the country of Piperfish. Muntaril is from the country of Clony. Muntaril produced the movie Skirtsicine with others. Monsterscar was one of the producers of the movie Featsaw. Monsterscar produced the movie Premercy with others. Monsterscar produced the movie Zalate with others. Teeplemole was one of the producers of the movie Featsaw. Teeplemole produced the movie Zalate with others. Muntaril produced the movie Premercy with others. Monsterscar wrote for the movie Premercy. Muntaril was one of the writers for the movie Zalate. Muntaril wrote for the movie Featsaw. Teeplemole wrote for the movie Featsaw. Monsterscar was one of the writers for the movie Zalate. Teeplemole was one of the writers for the movie Skirtsicine.
Q: Which awards were given to Zalate?
A: ["Hallowcock"]

Q: Which awards were given to Featsaw?
A: ["Zorgion"]

movie: Misgendery ; directed by: Wetherality. movie: Dewbar ; director: Gigabut. movie: Caudacite ; director: Lougerière. movie: Tayenne ; directed by: Lougerière. movie: Misgendery ; awarded: Microsouenesis. movie: Dewbar ; awarded: Erowid. movie: Tayenne ; awarded: Cockspit. movie: Caudacite ; award: Erowid. award: Aniconder ; winner: Wetherality. award: Aniconder ; winner: Lougerière. person: Gigabut ; award: Trifogation. movie: Dewbar ; release year: 1991. movie: Tayenne ; year: 2013. movie: Caudacite ; release year: 2008. movie: Misgendery ; year: 1991. Wetherality was an actor in the movie Dewbar. Gigabut was an actor in the movie Tayenne. Lougerière was an actor in the movie Tayenne. Lougerière acted in the movie Caudacite. Lougerière acted in the movie Misgendery. Gigabut was an actor in the movie Caudacite. Wetherality was an actor in the movie Misgendery. Wetherality was born in the year 1917. Lougerière was born in 1926. Gigabut was born in the year 1917. Gigabut grew up in the nation of Triclops. Lougerière is from the country of Tatkin. Wetherality grew up in the nation of Tatkin. Lougerière produced the movie Dewbar with others. Gigabut produced the movie Tayenne with others. Gigabut produced the movie Dewbar with others. Lougerière was one of the producers of the movie Misgendery. Wetherality was one of the producers of the movie Caudacite. Gigabut was one of the producers of the movie Caudacite. Wetherality produced the movie Misgendery with others. Wetherality produced the movie Tayenne with others. Wetherality wrote for the movie Tayenne. Gigabut wrote for the movie Misgendery. Lougerière was one of the writers for the movie Caudacite. Wetherality wrote for the movie Misgendery. Gigabut wrote for the movie Tayenne. Gigabut wrote for the movie Dewbar. Lougerière wrote for the movie Dewbar. Wetherality wrote for the movie Caudacite.
Q: Which movies were given the Erowid award?
A: ["Dewbar", "Caudacite"]

Q: Which awards were given to Wetherality?
A: ["Aniconder"]

movie: Pastillobox ; directed by: Firmline. movie: Clenestration ; directed by: Carblock. movie: Pestok ; directed by: Bioperatology. movie: Vitrilateral ; director: Bioperatology. movie: Vitrilateral ; award: Antipositive. movie: Clenestration ; awarded: Handt. movie: Pastillobox ; awarded: Handt. movie: Pestok ; awarded: Gutney. movie: Pestok ; writer: Firmline. movie: Clenestration ; written by: Carblock. movie: Pastillobox ; written by: Bioperatology. movie: Pestok ; writer: Bioperatology. movie: Clenestration ; written by: Firmline. movie: Vitrilateral ; writer: Bioperatology. movie: Pastillobox ; writer: Carblock. movie: Vitrilateral ; written by: Carblock. movie: Pestok ; release year: 1986. movie: Clenestration ; year: 1986. movie: Vitrilateral ; year: 1999. movie: Pastillobox ; release year: 1984. Carblock was an actor in the movie Pastillobox. Firmline was an actor in the movie Vitrilateral. Bioperatology was an actor in the movie Clenestration. Firmline acted in the movie Pastillobox. Carblock was an actor in the movie Clenestration. Bioperatology was an actor in the movie Pestok. Firmline was born in the year 1904. Bioperatology was born in the year 1935. Carblock was born in 1935. Carblock grew up in the nation of Knoppock. Firmline grew up in the nation of Tatkin. Bioperatology grew up in the nation of Tatkin. Bioperatology won the Modiparity award. Halfbill was awarded to Firmline. Halfbill was awarded to Carblock. Bioperatology was one of the producers of the movie Pestok. Bioperatology produced the movie Vitrilateral with others. Firmline produced the movie Pastillobox with others. Firmline produced the movie Clenestration with others. Carblock was one of the producers of the movie Pastillobox. Carblock produced the movie Vitrilateral with others. Carblock produced the movie Clenestration with others. Firmline was one of the producers of the movie Pestok.
Q: Who has been awarded the Modiparity award?
A: ["Bioperatology"]

Q: Which awards were given to Pestok?
A: ["Gutney"]

movie: Nohit ; director: Mimicocycle. movie: Noenometer ; director: Mimicocycle. movie: Tayenne ; directed by: Zayage. movie: Pneumodendron ; director: Sclerocybin. movie: Tayenne ; awarded: Goosehead. movie: Nohit ; awarded: Handt. movie: Pneumodendron ; award: Handt. movie: Noenometer ; awarded: Brownbeard. movie: Nohit ; writer: Mimicocycle. movie: Noenometer ; written by: Sclerocybin. movie: Tayenne ; writer: Sclerocybin. movie: Pneumodendron ; written by: Zayage. movie: Tayenne ; writer: Zayage. movie: Pneumodendron ; written by: Mimicocycle. movie: Noenometer ; release year: 1991. movie: Tayenne ; year: 2013. movie: Nohit ; year: 2005. movie: Pneumodendron ; year: 2005. Mimicocycle was an actor in the movie Tayenne. Zayage acted in the movie Pneumodendron. Zayage was an actor in the movie Nohit. Sclerocybin was an actor in the movie Nohit. Sclerocybin was an actor in the movie Tayenne. Mimicocycle was an actor in the movie Noenometer. Zayage was born in 1935. Sclerocybin was born in 1935. Mimicocycle was born in 1930. Mimicocycle is from the country of Calderita. Sclerocybin grew up in the nation of Calderita. Zayage is from the country of Obility. Quinion was awarded to Zayage. Fannyxist was awarded to Sclerocybin. Fannyxist was awarded to Mimicocycle. Mimicocycle produced the movie Nohit with others. Zayage was one of the producers of the movie Nohit. Sclerocybin was one of the producers of the movie Tayenne. Sclerocybin produced the movie Pneumodendron with others. Zayage produced the movie Pneumodendron with others. Mimicocycle was one of the producers of the movie Tayenne. Sclerocybin was one of the producers of the movie Noenometer. Zayage produced the movie Noenometer with others
Q: Which awards were given to Noenometer?
A: ["Brownbeard"]

Q: Which awards were given to Pneumodendron?
A: ["Handt"]
\end{lstlisting}
\lstinputlisting[title=\texttt{pos\_qa}]{prompts/commaqa/pos_qa.txt}
\lstinputlisting[title=\texttt{simp\_qa}]{prompts/commaqa/simp_qa.txt}

\subsubsection{COT}
\lstinputlisting[title=\texttt{COT}]{prompts/commaqa/cot.txt}

\subsection{Shorter Prompts for Smaller Context Windows}
\begin{lstlisting}[title=\texttt{qa(small)}]
movie: Premercy ; directed by: Muntaril. movie: Skirtsicine ; director: Teeplemole. movie: Featsaw ; directed by: Monsterscar. movie: Zalate ; director: Monsterscar. movie: Zalate ; awarded: Hallowcock. movie: Featsaw ; awarded: Zorgion. movie: Premercy ; award: Chowwurst. movie: Skirtsicine ; award: Hallowcock. award: Goatfly ; winner: Teeplemole. person: Monsterscar ; award: Glodome. person: Muntaril ; award: Goatfly. movie: Featsaw ; release year: 1973. movie: Zalate ; release year: 1964. movie: Skirtsicine ; release year: 1973. movie: Premercy ; year: 1961. Teeplemole was an actor in the movie Skirtsicine. Muntaril was an actor in the movie Skirtsicine. Monsterscar was an actor in the movie Premercy. Muntaril was an actor in the movie Featsaw. Teeplemole was an actor in the movie Zalate. Muntaril was born in the year 1910. Teeplemole was born in 1910. Monsterscar was born in 1942. Teeplemole is from the country of Piperfish. Monsterscar is from the country of Piperfish. Muntaril is from the country of Clony. Muntaril produced the movie Skirtsicine with others. Monsterscar was one of the producers of the movie Featsaw. Monsterscar produced the movie Premercy with others. Monsterscar produced the movie Zalate with others. Teeplemole was one of the producers of the movie Featsaw. Teeplemole produced the movie Zalate with others. Muntaril produced the movie Premercy with others. Monsterscar wrote for the movie Premercy. Muntaril was one of the writers for the movie Zalate. Muntaril wrote for the movie Featsaw. Teeplemole wrote for the movie Featsaw. Monsterscar was one of the writers for the movie Zalate. Teeplemole was one of the writers for the movie Skirtsicine.
Q: Which awards were given to Zalate?
A: ["Hallowcock"]

Q: For which movies was Muntaril the producer?
A: ["Premercy]

Q: Who are the actors in the movie Premercy?
A: ["Monsterscar"]

movie: Nilitude ; director: Monsterscar. movie: Dewbar ; directed by: Metatoun. movie: Warpstone ; directed by: Gastrat. movie: Partnershipmaker ; director: Metatoun. movie: Dewbar ; award: Tachychronograph. movie: Partnershipmaker ; awarded: Tachychronograph. movie: Nilitude ; award: Paleodactyl. movie: Warpstone ; award: Sabonade. person: Gastrat ; award: Trifogation. award: Polyquadrase ; winner: Monsterscar. award: Trifogation ; winner: Metatoun. movie: Warpstone ; release year: 1956. movie: Dewbar ; release year: 1984. movie: Nilitude ; year: 1984. movie: Partnershipmaker ; year: 1962. Gastrat was an actor in the movie Partnershipmaker. Metatoun was an actor in the movie Partnershipmaker. Metatoun was an actor in the movie Nilitude. Gastrat acted in the movie Nilitude. Monsterscar was an actor in the movie Dewbar. Gastrat acted in the movie Warpstone. Metatoun acted in the movie Warpstone. Metatoun was born in 1939. Gastrat was born in the year 1933. Monsterscar was born in 1933. Metatoun grew up in the nation of Moulole. Gastrat is from the country of Stridery. Monsterscar grew up in the nation of Moulole. Monsterscar produced the movie Nilitude with others. Monsterscar was one of the producers of the movie Warpstone. Metatoun was one of the producers of the movie Warpstone. Gastrat was one of the producers of the movie Nilitude. Metatoun produced the movie Partnershipmaker with others. Metatoun produced the movie Dewbar with others. Monsterscar was one of the producers of the movie Partnershipmaker. Gastrat produced the movie Dewbar with others. Metatoun wrote for the movie Partnershipmaker. Gastrat wrote for the movie Warpstone. Gastrat was one of the writers for the movie Dewbar. Monsterscar was one of the writers for the movie Nilitude. Metatoun wrote for the movie Warpstone.
Q: Which movies has Gastrat been an actor in?
A: ["Partnershipmaker", "Nilitude", "Warpstone"]

Q: Who is from the country Stridery?
A: ["Gastrat"]

Q: Which movies were given the Tachychronograph award?
A: ["Dewbar", "Partnershipmaker"]
\end{lstlisting}
\lstinputlisting[title=\texttt{aw\_qa(small)}]{prompts/commaqa/aw_qa_small.txt}
\lstinputlisting[title=\texttt{pos\_qa(small)}]{prompts/commaqa/pos_qa_small.txt}
\lstinputlisting[title=\texttt{simp\_qa(small)}]{prompts/commaqa/simp_qa_small.txt}
\lstinputlisting[title=\texttt{COT(small)}]{prompts/commaqa/cot_small.txt}


\subsection{Math QA}
The decomposer here deterministically calls \texttt{cot} to generate the CoT and then calls \texttt{gpt\_ans} to extract the answer.

\lstinputlisting[title=\texttt{cot}]{prompts/mathqa/cot.txt}
\begin{lstlisting}[title=\texttt{gpt\_ans}]
Q: There are 15 trees originally. Then there were 21 trees after some more were planted. So there must have been 21 - 15 = 6 trees planted.
A: 6

Q: There are originally 3 cars. 2 more cars arrive. 3 + 2 = 5. The answer is 5.
A: 5

Q: Originally, Leah had 32 chocolates. Her sister had 42. So in total they had 32 + 42 = 74. After eating 35, they had 74 - 35 = 39. The answer is 39.
A: 39

Q: Jason started with 20 lollipops. Then he had 12 after giving some to Denny. So he gave Denny 20 - 12 = 8 apples.
A: 8

Q: Shawn started with 5 toys. If he got 2 toys each from his mom and dad, then that is 4 more toys. 5 + 4 = 9. The answer is 9.
A: 9

Q: There were originally 9 computers. For each of 4 days, 5 more computers were added. So 5 * 4 = 20 computers were added. 9 + 20 is 29. The answer is 29.
A: 29

Q: Michael started with 58 golf balls. After losing 23 on tuesday, he had 58 - 23 = 35. After losing 2 more, he had 35 - 2 = 33 golf balls.
A: 33

Q: Olivia had 23 dollars. 5 bagels for 3 dollars each will be 5 x 3 = 15 dollars. So she has 23 - 15 = 8 dollars left.
A: 8
\end{lstlisting}

\subsection{Open Domain QA}
The prompts in this section implement Decomposed Prompting approach to open-domain multihop QA. For brevity we've included prompts for 5 of 20 randomly sampled questions. The full prompts are attached with the submission and will also be released with the code. Note that we selected a set of 100 questions from the development set to tune the hyperparameter (number of paragraphs to retrieve for all of the retrieval-based approaches).

\lstinputlisting[title=\texttt{hotpotqa: decomp}]{prompts/odqa/hotpotqa__decomp.txt}
\lstinputlisting[title=\texttt{hotpotqa: retrieve\_odqa}]{prompts/odqa/hotpotqa__retrieve_odqa.txt}
\lstinputlisting[title=\texttt{hotpotqa: singlehop\_titleqa}]{prompts/odqa/hotpotqa__singlehop_titleqa.txt}
\lstinputlisting[title=\texttt{hotpotqa: multihop\_titleqa (direct)}]{prompts/odqa/hotpotqa__multihop_titleqa_direct.txt}
\lstinputlisting[title=\texttt{hotpotqa: multihop\_titleqa (cot)}]{prompts/odqa/hotpotqa__multihop_titleqa_cot.txt}

\lstinputlisting[title=\texttt{2wikimultihopqa: decomp}]{prompts/odqa/2wikimultihopqa__decomp.txt}
\lstinputlisting[title=\texttt{2wikimultihopqa: retrieve\_odqa}]{prompts/odqa/2wikimultihopqa__retrieve_odqa.txt}
\lstinputlisting[title=\texttt{2wikimultihopqa: singlehop\_titleqa}]{prompts/odqa/2wikimultihopqa__singlehop_titleqa.txt}
\lstinputlisting[title=\texttt{2wikimultihopqa: multihop\_titleqa (direct)}]{prompts/odqa/2wikimultihopqa__multihop_titleqa_direct.txt}
\lstinputlisting[title=\texttt{2wikimultihopqa: multihop\_titleqa (cot)}]{prompts/odqa/2wikimultihopqa__multihop_titleqa_cot.txt}

\lstinputlisting[title=\texttt{musique\_ans: decomp}]{prompts/odqa/musique_ans__decomp.txt}
\lstinputlisting[title=\texttt{musique\_ans: retrieve\_odqa}]{prompts/odqa/musique_ans__retrieve_odqa.txt}
\lstinputlisting[title=\texttt{musique\_ans: singlehop\_titleqa}]{prompts/odqa/musique_ans__singlehop_titleqa.txt}
\lstinputlisting[title=\texttt{musique\_ans: multihop\_titleqa (direct)}]{prompts/odqa/musique_ans__multihop_titleqa_direct.txt}
\lstinputlisting[title=\texttt{musique\_ans: multihop\_titleqa (cot)}]{prompts/odqa/musique_ans__multihop_titleqa_cot.txt}

\end{document}